\documentclass[jair,twoside,11pt]{article}
\usepackage{jair, rawfonts}
\usepackage[utf8]{inputenc} %
\usepackage{hyperref}       %
\usepackage{url}            %
\usepackage{booktabs}       %
\usepackage{amsfonts}       %
\usepackage{nicefrac}       %
\usepackage{microtype}      %
\usepackage{amsmath, amsthm}
\usepackage{hyperref}
\usepackage[capitalise,nameinlink,noabbrev]{cleveref}
\usepackage{color}
\usepackage{todonotes}
\usepackage[round]{natbib}
\definecolor{cb-blue-light} {RGB}{182, 219, 255}
\definecolor{cb-blue-green} {RGB}{  0,  073,  073}
\definecolor{cb-purple}     {RGB}{ 73,   0, 146}
\definecolor{cb-rose}       {RGB}{255, 109, 182}
 \hypersetup{         
     colorlinks=true,
     linkcolor=cb-rose,
     citecolor=cb-purple,
     urlcolor=cb-blue-light,
}
\newcommand{\indep}{\perp \!\!\! \perp}

\usepackage{bm}
\usepackage{enumerate}
\usepackage[framemethod=TikZ]{mdframed}
\newenvironment{enumerate*}%
  {\begin{enumerate}%
    \vskip 0.1in%
    \setlength{\itemsep}{0pt}%
    \setlength{\parskip}{0pt}}%
  {\end{enumerate}%
   \vskip -0.1in}

\definecolor{dblue}{RGB}{98, 140, 190}
\definecolor{dlblue}{RGB}{216, 235, 255}
\definecolor{dgreen}{RGB}{124, 155, 127}
\definecolor{dpink}{RGB}{207, 166, 208}
\definecolor{dyellow}{RGB}{255, 248, 199}
\definecolor{dgray}{RGB}{46, 49, 49}

\newcommand{\durl}[1]{\textcolor{dblue}{\underline{\url{#1}}}}

\newcounter{KDefCounter}
\newcommand{\ddef}[2]
{
\vspace{1mm}
{\bf Definition \theKDefCounter} (#1): {\it #2}
\stepcounter{KDefCounter}
}

\newtheorem{proposition}{Proposition}

\begin{document}
\firstpageno{1}

\title{Towards Continual Reinforcement Learning: \\
A Review and Perspectives}

\author{\name Khimya Khetarpal\thanks{Equal contribution.} \email khimya.khetarpal@mail.mcgill.ca \\
       \addr Mila, McGill University
       \AND
       \name Matthew Riemer\footnotemark[1] \email mdriemer@us.ibm.com \\
       \addr Mila, Universit\'e de Montr\'eal, IBM Research
       \AND
       \name Irina Rish \email irina.rish@mila.quebec\\
       \addr Mila, Universit\'e de Montr\'eal
       \AND
       \name  Doina Precup \email dprecup@cs.mcgill.ca\\
       \addr Mila, McGill University, DeepMind
   }

\maketitle

\begin{abstract}
In this article, we aim to provide a literature review of different formulations and approaches to continual reinforcement learning (RL), also known as lifelong or non-stationary RL. We begin by discussing our perspective on why RL is a natural fit for studying continual learning. We then provide a taxonomy of different continual RL formulations by mathematically characterizing two key properties of non-stationarity, namely, the scope and driver non-stationarity. This offers a unified view of various formulations. Next, we review and present a taxonomy of continual RL approaches. We go on to discuss evaluation of continual RL agents, providing an overview of benchmarks used in the literature and important metrics for understanding agent performance. Finally, we highlight open problems and challenges in bridging the gap between the current state of continual RL and findings in neuroscience. While still in its early days, the study of continual RL has the promise to develop better incremental reinforcement learners that can function in increasingly realistic applications where non-stationarity plays a vital role. These include applications such as those in the fields of healthcare, education, logistics, and robotics.\footnote{This work has been accepted by the Journal of Artificial Intelligence Research (JAIR).} 
\end{abstract}

\section{Introduction}
Recent advances in deep RL have demonstrated superhuman performance by artificially intelligent (AI) agents on a variety of impressive tasks. However, current approaches for achieving these results center around an agent that primarily learns how to master a narrow task of interest. Meanwhile, untrained agents often need to play far more %
of these games over their lifetime than their human competition and even after doing so, lack the ability to generalize to new variations even for simple RL problems \citep{bengio2020interference}. In contrast, humans have a remarkable ability to continually learn and adapt to new scenarios over the duration of their lifetime. This ability is referred to as continual learning. {\em Continual learning (CL)} is the constant and incremental development of increasingly complex behaviors. This includes the process of building complicated behaviors on top of those already developed \citep{ring1997child} while being able to reapply, adapt, and generalize previously learned abilities to new situations. CL is a rapidly growing area of modern machine learning and particularly so for the study of deep learning. It is also closely related to settings such as \textit{lifelong learning}, \textit{online learning} or \textit{never-ending learning}. In this paper, we are concerned with {\em continual RL}. This is a natural fit, as RL inherently provides an agent-environment interaction paradigm amenable to studying the topic of learning in a continual fashion. 

A continual learner can be seen as an autonomous agent learning over an endless stream of tasks, which has the following desiderata in the context of learning: 1) it can learn online, 2) it learns behaviors or skills while solving presented tasks, 3) learning is task agnostic, 4) it learns incrementally with no fixed training set, 5) it learns behaviors that can be built upon later, 6) it retains previously learned abilities i.e. it minimizes catastrophic forgetting and interference, and 7) it adapts efficiently to changes experienced over time and recovers quickly. In its most ambitious form, CL occurs at every moment with no bounded set of tasks or data sets and no clearly presented boundaries between tasks. The learning agent should be able to transfer and adapt what it has learned from previous experiences, data, or tasks to new situations and make use of more recent experiences to improve performance on capabilities learned earlier.

With the rapidly growing interest in continual learning research, there have been extensive reviews of a large body of work in supervised continual learning such as \citep{parisi2019continual}, \citep{de2019continual}, \citep{mundt2020wholistic} and \citep{HADSELL20201028}, %
Most of this work considers the \textit{task incremental learning setting}. In this setting, each task is received with its training data in the form of labelled samples of inputs and desired outputs $( \mathcal{X}, \mathcal{Y})$ that are randomly drawn from a distribution $\mathcal{D}$. Here, the goal is statistical risk minimization on all seen tasks given limited or no access to the data from previous tasks after initial learning. In contrast, continual RL involves a sequential decision making problem over a stream of tasks where each task can be considered a stationary \textit{Markov Decision Process} (MDP)~\citep{puterman1994markov}. 

\paragraph{Key Contributions.} Due to the generality of the continual learning problem and the struggle to define its scope, researchers have often interchangeably used the terms \textit{multi-task learning}, \textit{lifelong learning}, and \textit{continual learning} in the field of RL. One of the primary goals of this work is to provide a concrete taxonomy of the different formulations and approaches under the broad umbrella term continual RL. To this end, the key contributions of this work concern: 1) a taxonomy and review of relevant problem formulations, 2) a taxonomy and review of families of approaches considered, and 3) a discussion of evaluation metrics %
for assessing continual RL agents and how relevant benchmarks %
can be used to generate non-stationarity during learning. Finally, we discuss connections to neuroscience as well as %
perspectives on challenges and open problems in the field.

\paragraph{Scope and Overview of the Survey.}
In this work, we discuss the literature which addresses different perspectives on continual RL. This includes multi-task learning, meta-learning, never-ending learning, non-stationary RL, and lifelong learning in the context of RL, explicitly. We limit our scope to the aforementioned topics. While there are several related topics such as transfer learning, representation learning, domain adaptation, and domain randomization, %
we do not cover them in detail in our survey. %
We first introduce the RL paradigm (Sec.~\ref{sec:background}) and highlight research directions related to the study of continual RL in Sec.~\ref{sec:spectrum}. We then proceed to discuss why reinforcement learning as a paradigm is a natural fit for studying continual learning in Sec.~\ref{sec:posingrl}. To this end, we present a broad taxonomy of continual RL formalism and approaches in Sec.~\ref{sec:taxonomyformalism} and Sec.~\ref{sec:taxonomyapproaches}, respectively. We then consider current and potential directions for evaluation of continual RL agents in Sec.~\ref{sec:evaluation}. Looking to the future, we conclude by presenting connections to findings in neuroscience and by discussing perspectives on challenges and open problems in the field in Sec.~\ref{sec:lookingforward}.

\section{Background}
\label{sec:background}

\textbf{Notation:} In this survey capital letters are used for random variables, while lower case letters are used for the values of random variables and for scalar functions. For consistency with prior literature, we largely follow the notation of \citep{Sutton98}. 

Typically, we formalize reinforcement learning based on a finite, discrete-time MDP \citep{puterman1994markov, Sutton98}, which is a tuple $ M = \langle {\cal S}, {\cal A}, p, r, \gamma \rangle $, where ${\cal S}$ is the set of states, ${\cal A}$ is the  set of actions, $r: {\cal S} \times {\cal A}\rightarrow \mathbb{R}$ is the reward function, $p:{\cal S} \times {\cal A} \rightarrow Dist({\cal S})$ is the environment transition probability function, and $\gamma \in [0,1)$ is the discount factor. At each time step, the learning agent perceives a state $s \in {\cal S}$, takes an action $a \in {\cal A}$ drawn from a policy $\pi : {\cal S} \times {\cal A} \rightarrow [0,1]$ with internal parameters $\theta \in \Theta$, and with probability $p(s'|s,a)$ enters next state $s'$, receiving a numerical reward $r(s,a)$ from the environment. It is also possible that the environment is a partially observable POMDP environment \citep{kaelbling1993learning}. In this case, the environment also consists of a (potentially stochastic) function that generates observations $o$ from the current state with probability $x(o|s)$. In these environments, agents must generate a belief about the current state based on their history of interactions and perform RL based on this belief. As such, we will mostly focus on reasoning with respect to the true environment state throughout this survey as extensions to partially observable settings are straightforward.  %

The environment's transition dynamics can be modeled by the one-step \textit{state-transition probabilities}, 
\begin{equation}
\label{eq:state-transition-probabilities}
    p(s'|s,a) \doteq P_{ss'}^{a} = Pr(S_{t+1}=s' | S_t=s, A_t=a)  %
\end{equation}
and one-step expected rewards,
\begin{equation}
\label{eq:one-step-expected-reward}
    r(s,a) = R_{s}^{a} = E[R_{t+1} | S_t=s, A_t=a] %
\end{equation}
for all $s, s'\in S$ and $a \in A$. $P_{ss'}^{a}$ and $R_{s}^{a}$ together form the one-step model of the environment.

The goal of the agent is to maximize the expected value of the accumulated discounted reward from time-step $t$. More precisely, the \textit{return} $G_t$ obtained from time step $t$ is defined as:
\begin{equation}
\label{eq:return}
    G_t \doteq R_{t+1} + \gamma R_{t+2} + \gamma ^{2} R_{t+3} + ... = \sum_{k=0} ^{\infty} \gamma^{k} R_{t+k+1}
\end{equation}
where $ 0 \leq \gamma < 1$ is the \textit{discount factor}. The agent's behavior is determined by its \textit{policy} (stochastic and stationary), a mapping from states to probabilities of taking each of the admissible primitive actions, $\pi: S \times A \rightarrow[0,1]$. The value of being in a state is determined by the \textit{state-value function} $v_{\pi}(s)$, defined as the expected return starting from state $s$, and then following policy $\pi$ is defined as:
\begin{equation*}
    v_\pi(s) =  \mathbb{E}_{\pi}\bigg[ G_t \bigg| S_t=s \bigg] = \sum_a \pi(a|s) \bigg[ r(s,a) + \gamma \sum_{s'} p(s'|s,a) v_{\pi}(s') \bigg]
\end{equation*}
Analogous to the state value function, the \textit{action-value function} $q_\pi(s,a)$ is defined as:
\begin{equation*}
q_\pi(s,a) = \mathbb{E}_{\pi}\bigg[ G_t \bigg| S_t=s, A_t=a \bigg] =  r(s,a) + \gamma \sum_{s'} p(s'|s,a) v_{\pi}(s')
\end{equation*}

For a finite MDP, there exists at least one deterministic policy, that is better than or equal to all other policies. This is an optimal policy $\pi^{*}$. The optimal policy $\pi^{*}$ achieves the optimal state-value function or the optimal action-value function. The \textit{optimal state-value function} $v_{*}(s)$ is the maximum value function over the class of stationary policies as defined in equation \eqref{optimal-state-value-function}. Similarly, the \textit{optimal action-value function} $q_{*}(s,a)$ is the maximum action-value-function over all policies as defined in equation \eqref{optimal-action-value-function}. 
\begin{equation}
\label{optimal-state-value-function}
    v_{*}(s) = \max_{\pi} v_{\pi}(s)
\end{equation}

\begin{equation}
\label{optimal-action-value-function}
    q_{*}(s,a) = \max_{\pi} q_{\pi}(s,a)
\end{equation}

RL algorithms can be classified as \textit{model-free} or \textit{model-based} in nature. Knowing or estimating a model of the environment, and using this model to compute value functions and learn policies is called \textit{planning} in RL. Algorithms such as policy-evaluation exploit the iterative Bellman expectation backup to find optimal solutions for prediction problems. However, in most practical situations, the model of the world is rarely known and many practical algorithms fall in the \textit{model-free} regime.

\subsection{Defining Tasks in RL}
\label{sec:taskdefinition}

\begin{figure}[h!]
   \vskip 0.2in
  \begin{center}
  \centerline{\includegraphics[width=0.9\linewidth]{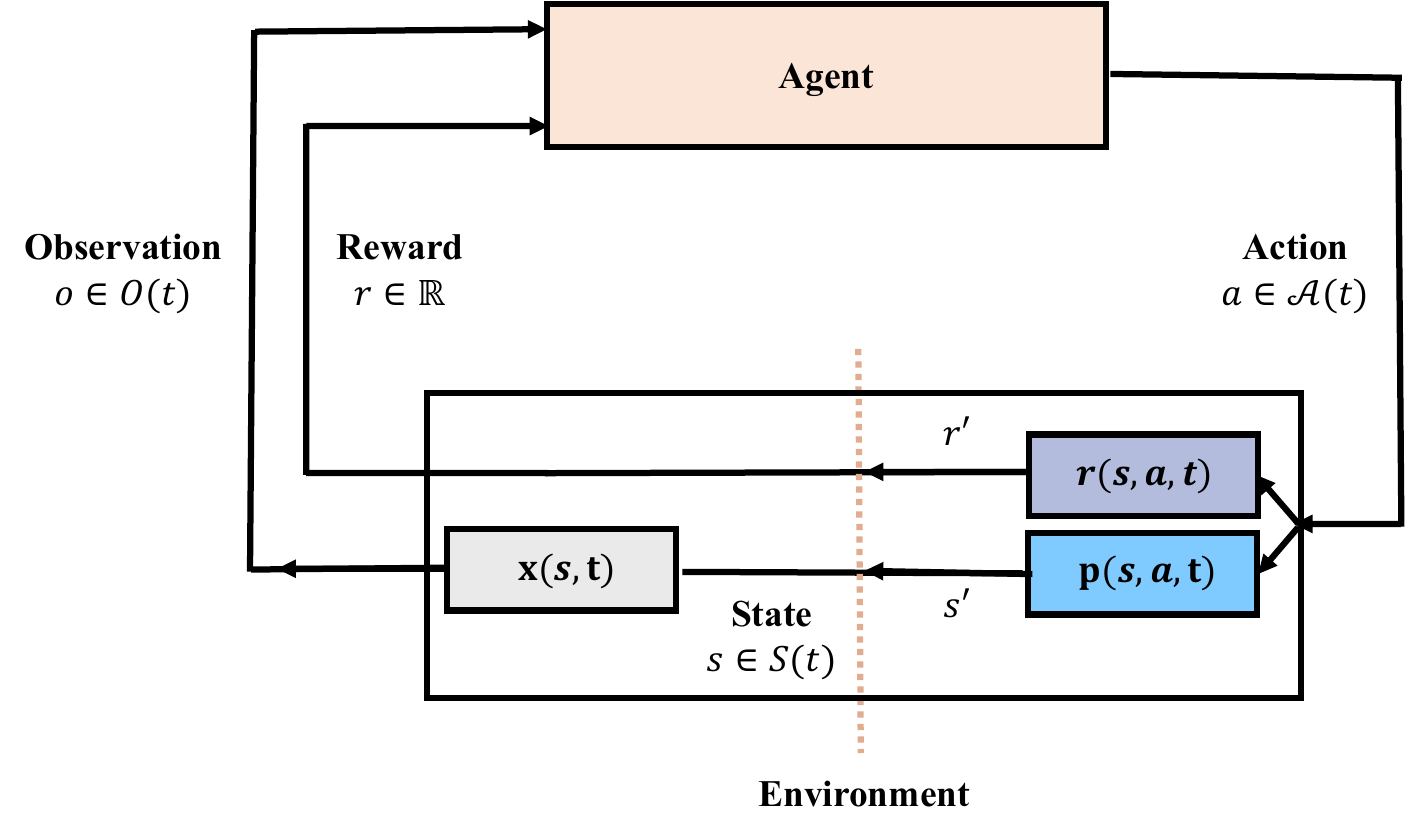}}
   \caption{\textbf{Agent-Environment Interaction with Potentially Time Dependent Environment Components.} Extending Figure 3.1 of \citep{Sutton98} to highlight the agent-environment interaction in continual reinforcement learning.}
   \label{agent-env-interface}
   \end{center}
  \vskip -0.2in
\end{figure}

In the RL literature the concepts of tasks and non-stationarity have been defined in many different ways depending on the context. As such, this is a point of confusion in the literature that we hope to provide some clarity on here. From our perspective, there are two primary views that have been taken in past work. In the first view, we consider that actual components of the RL environment may exhibit some time dependence. We will position most of our survey in terms of this view as it is the most common perspective taken in the continual learning literature to date. However, another valid perspective to take is that the underlying physics of the world fundamentally exhibit stationary dynamics and that perceived non-stationarity is really only a consequence of unobserved phenomena that result in changed dynamics from an agent's own (potentially ignorant) perspective. We will take this view only when discussing approaches for \textit{context detection} in Sec. \ref{sec:contextdetection} or approaches for multi-agent RL as this view is key to the underlying theory behind these techniques. 

\textbf{Non-stationary Function View:} We highlight the view that fundamental components of the RL environment may exhibit time dependence in Figure \ref{agent-env-interface}. Indeed, in the most extreme case, it is possible that the transition function, reward function, observation function, and action space may all depend on time. In this setting, for the purposes of this paper, we will define a task $z$ as constituting a stationary MDP $M^{(z)} = \langle \mathcal{S}^{(z)}, \mathcal{A}^{(z)}, p^{(z)}, r^{(z)}, \gamma^{(z)} \rangle$ with initial state distribution $p_{0}^{(z)}$. This implies that there is a discrete set of tasks. However, in the most extreme case, this set may be of infinite size where no task is ever visited for more than a single time step. In principle, it is also possible that MDPs may vary as a continuous function of time. That said, this distinction is only relevant for RL in continuous time MDPs, which is beyond the scope of this paper. 

\textbf{Partially Observable View:} Another view on non-stationarity is that it is not a feature of the environment itself, but rather only the agent's perspective in that environment. This viewpoint certainly appears realistic when we make comparisons to human learning. It appears that most non-stationarity that humans experience in their lives is the result of interacting in a huge environment with a massive number of agents who are changing their behaviors for possibly unknown reasons over time. When we take this view in the context of RL, a task is defined as an unobserved component of the state that an agent must develop beliefs about to achieve optimal performance. Indeed, as explained in \citep{xie2020deep}, non-stationarity chiefly adds a non-Markovian aspect to the learning setting that the POMDP framework helps directly address. 

While these two views of tasks and non-stationarity may seem contradictory on the surface, we believe that they actually serve as complementary perspectives on the same problem. The non-stationary function view describes the agents perspective on the problem and is easier to reconcile with similar formulations in the context of supervised learning and bandit learning. Meanwhile, the partially observable view makes it easier to formalize convergence analysis. Moreover, the partially observable view puts emphasis on different aspects of the problem by focusing less on non-stationarity and more on related complications like non-Markovian observations and non-ergodic environments (or even ergodic environments with very high mixing times \citep{polynomial}). 

\subsection{A Spectrum of Learning Settings}
\label{sec:spectrum}

So far we have discussed the basics of RL in a stationary environment. However, in general, it is possible for each component of our environment to be non-stationary. In Figure \ref{agent-env-interface} we illustrate the RL setup where the environment includes explicit time dependence in each component. This represents the most ambitious formalization of continual RL (as we will discuss in more detail in Sec. \ref{sec:taxonomyformalism}). Indeed, the agent must deal with potential non-stationarity in state transitions, the reward function, the way observations are produced, and even the availability of actions over time. As this setting presents a number of serious challenges, much of the work in the community has focused on less ambitious variations of the problem that are more targeted to highlight particular aspects of the difficulty encountered by agents as they continual learn. As such, in this section, we will highlight a spectrum of research directions of particular relevance to the study of continual RL.

\begin{figure}[h!]
  \begin{center}
  \includegraphics[width=1.0\textwidth]{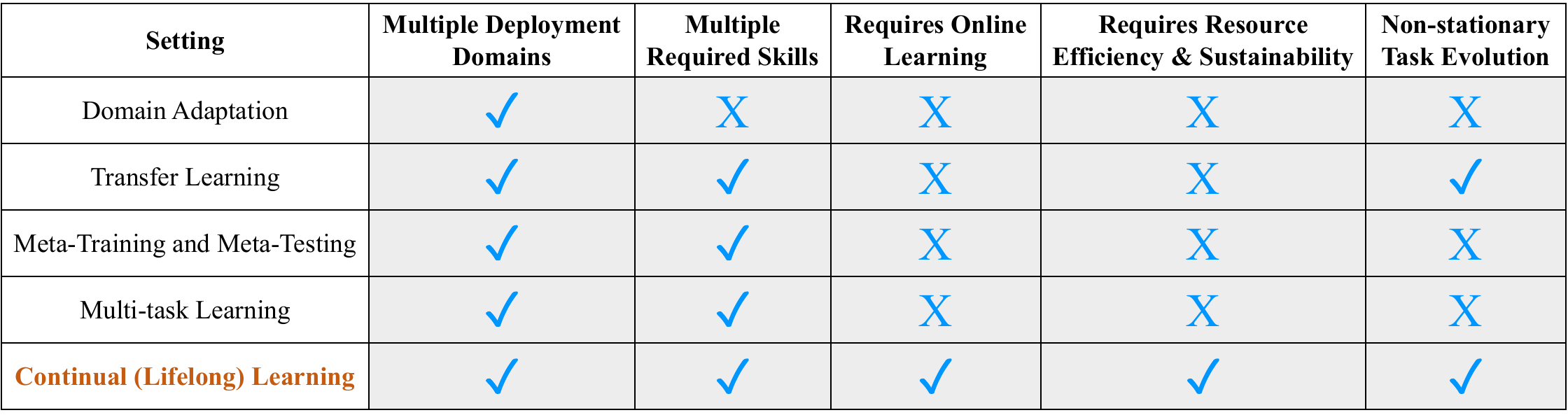}
  \caption{\textbf{A Spectrum of Learning Settings}: For each setting we consider whether they typically involve multiple domains, multiple skills, online learning, resource efficiency/sustainability and a non-stationary evolution of the task distribution.}
   \label{fig:settings}
   \end{center}
   \vspace{-20pt}
\end{figure}

\textbf{Domain Adaptation:} As detailed in Figure \ref{fig:settings}, domain adaptation is the process of adapting a policy for a specific skill to a new domain. It generally involves building a separate policy for each domain and typically does not require algorithms to address environment non-stationary during training. A natural use case for domain adaptation in the context of reinforcement learning is \textit{sim2real} transfer when an agent is trained in a simulation environment and then adapted as a result of interaction in the real world. Domain randomization is a related approach that deals with a transfer type of setting. The goal here is to be able to generalize learning from source domains to target domains. We refer the reader to \citep{joshtobin2019} for a comprehensive discussion on domain randomization and related topics such as domain adaptation.

\textbf{Transfer in RL:} Learning each task from scratch may require a huge amount of data to achieve adequate performance. Additionally, learning from scratch is computationally expensive and intractable for large scale problems such as everyday robotics. However, to learn about multiple potentially diverse tasks with limited data is also a challenging problem. A large body of work concerned with an agent's performance on more than one task has extensively studied the topic of {\em transfer learning}: training on \textit{source} tasks to perform efficient policy modifications during training on a single \textit{target} task drawn from a distribution of related tasks. We refer the reader to \citep{taylor2009transfer} for an extensive review of work focused around transfer learning in RL. As highlighted in Figure \ref{fig:settings}, transfer learning settings generally include multiple domains and skills that have to be learned in the presence of non-stationarity. However, a key distinction with other settings of more interest to the field of continual RL is that in transfer learning settings it is generally assumed that a separate policy is learned for each task and that task boundaries are given. In transfer learning, the learning process is generally broken up into distinct phases such as \textit{pre-training} and \textit{fine-tuning}.

\textbf{Meta-Training and Meta-Testing:} A related setting is the \textit{meta-training} and \textit{meta-testing} protocol common in the meta-learning literature. In this setting, an agent first performs \textit{meta-training} about how to learn to generalize efficiently on a distribution of tasks and this meta-learning model is transferred to a \textit{meta-testing} distribution of tasks where the eventual performance of the policy of each task is used as a basis for comparison. As indicated in Figure \ref{fig:settings}, in some sense this meta-learning protocol results in an easier optimization process than we see in traditional transfer learning. This is because \textit{meta-training} consists of drawing tasks to learn on from a stationary distribution and that distribution should theoretically be approximately stationary during \textit{meta-testing} as well. As a result, the evolution of tasks is not really non-stationary as in more generic transfer learning. See Sec. \ref{sec:learningtoadapt} for a more in depth discussion of this setting.

\textbf{Multi-task RL:} More closely related to the aforementioned task incremental learning setting, is \textit{multi-task reinforcement learning}. 
In a commonly used formulation of multi-task RL,  the agent is required to learn a series of sequential decision making tasks $M^{(1)},  . . . , M^{(z_{max})}$ over its lifetime. %
The agent will learn the tasks consecutively, potentially acquiring multiple trajectories within each task before moving to the next. It is a common assumption that these tasks may be interleaved i.e. the agent might revisit earlier tasks, but the agent does not control the order of tasks. This setting has often been studied for {\em online reinforcement learning} \citep{wilson2007multi, ammar2014online}. Alternatively the tasks do not necessarily arrive in a sequential fashion, and learning might not be considered in a fully online setting. In contrast, data could be generated by many different behaviour policies and be made available in a batch generated beforehand in the {\em offline reinforcement learning} setting. We refer the reader to \citep{levine2020offline} for a tutorial and review of the offline setting. In the well posed case of multi-task reinforcement learning, the agent's overall objective is to maximize performance across the distribution of tasks being considered. It is desired that such an agent also perform well on out-of-distribution data usually not seen during training and generalize to similar or related tasks. How this is achieved could vary from learning a single universal policy whilst maximizing the expected accumulated average return on all tasks, to learning a set of optimal policies for each task separately, to learning a set of shared skills leveraging a meta-controller. 

\textbf{Continual (Lifelong) RL:} A prime challenge faced by continual RL agents is to be able to retrieve relevant information from a massive sensory data stream. A common approach entails compressing information in one way or another such as discovering information bottleneck states, subgoals, or state abstractions to name a few. Additionally, a continual learner has no direct access to all previous experiences and its memory is often limited. For such an agent, it is essential to properly assign credit to key events over the course of its lifetime. This difficulty is commonly known as the \textit{credit-assignment} problem. Additionally, while successful continual learning agents must have resistance to catastrophic forgetting, it is not immediately clear if agents must perform well on all previously seen tasks. As we note in Figure \ref{fig:settings}, continual learning adds the concern of learning over a non-stationary task distribution to the complication of learning a policy over all tasks. For instance, someone who learns to play tennis at a young age may not necessarily perform well on this previously seen task in later stages of life if it has not been rehearsed. As such, quick adaptation and building on relevant previously learned behaviors are also central to the study of continual RL.

\subsection{Important Related Topics}

When it comes to actually achieving good continual RL performance, it is not possible to totally disentangle the study of continual learning from other very important fields of study in RL. 

\textbf{Representation Learning:} Learning good representations with minimal overlap is fundamental to much of the work related to multi-task deep RL. Considering representation learning is not unique to continual RL and is a common concern across literature on supervised continual learning, curriculum learning, transfer learning, multi-task learning, and more generally work towards broader AI. We refer the reader to discussions on encoder based lifelong learning by \citep{de2019continual}.

\textbf{Generalization in RL:}
Due to the inherent difficulty of training and testing on the same environment in deep reinforcement learning, several efforts have been made in studying the generalization abilities of RL agents. Generalization in RL has been investigated by creating different game modes \citep{farebrother2018generalization} and game levels \citep{nichol2018gotta} for training and testing, by exposing internal parameters of various classic environments \citep{packer2018assessing}, and by procedurally generating environments \citep{justesen2018illuminating, zhang2018study, cobbe2018quantifying}. There has indeed been surprising evidence to the extent that RL agents tend to overfit even in simple settings \citep{bengio2020interference}. Recent theoretical work \citep{du2019good} suggests that perhaps for a class of tree-like MDPs, generalization might even be improbable. While robust generalization is a central capability for effective continual RL, most of the literature on generalization in RL studies it within the scope of a simple transfer learning setting. In this way, it is often possible to study generalization without conflating its properties with the optimization difficulties specific to continual RL. 

\section{RL: A Natural Fit For Studying Continual Learning}
\label{sec:posingrl}
It has been well known for decades that the primary challenge for neural networks when learning over a non-stationary stream of data is balancing the \textit{stability-plasticity dilemma} \citep{StabilityPlasticity}. This dilemma highlights the tension between prioritizing recent experiences and past experiences when training neural networks. A common failure case is the so called \textit{catastrophic forgetting} problem \citep{CF}, where the network adapts to recent experiences while significantly deteriorating its capabilities on past experiences. However, in a certain sense, the formalism of RL in continuing environments actually provides us with means of directly studying the \textit{stability-plasticity dilemma} as we will highlight in this section. 

For RL in a continuing environment, we define an objective function $J_{\text{continuing}}$ which is to learn a policy $\pi$ that maximizes its value function $v_\pi(s)$. We seek to maximize this infinite horizon objective function at each point in time:

\begin{equation} \label{infinitehorizonobjective}
\begin{split}
J_{\text{continuing}}(\pi) = v_\pi(s) &= \mathbb{E}_{\pi} \bigg[ \sum_{k=0}^\infty \gamma^k R_{t+k} \bigg| S_t =s \bigg] 
\end{split}
\end{equation} 

Here our discounted objective with respect to $\pi$ is to maximize the expected long-term discounted returns of this policy in the current state $s$. %
So, the objective we would like to maximize does not just concern itself with the current state, but rather the full expected future distribution of states as well. Some proportion of the expected future distribution of states is likely to be similar to states from the past. As such, this provides a general paradigm for addressing the prioritization problem of the \textit{stability-plasticity dilemma}. We care about the present and the past proportionally to how representative these distributions will be of the future. However, because the future is generally unknown, we will have to use our expectations about the future to guide our prioritization instead. 

As RL in continuing environments is clearly a challenging problem, the vast majority of work has focused on easier learning problems such as RL in episodic environments and supervised learning. However, when we introduce non-stationary environment dynamics to either the episodic or supervised settings, naive approaches can be shown to produce myopic biased updates that are overly focused on the current experience distribution rather than the expected future distribution.  This is because non-stationary dynamics undermine the assumptions of many popular algorithms for these settings such as policy gradient based approaches in episodic reinforcement learning and stochastic gradient descent (SGD) in supervised learning. However, even settings with changing time correlated dynamics can be modeled in terms of the formulation of RL in continuing environments. This perspective can be illuminating and shed light on the pervasive \textit{catastrophic forgetting} effects experienced by popular approaches when applied to non-stationary settings.

\paragraph{Catastrophic Forgetting in Continual Episodic Reinforcement Learning} 
Much of the progress in deep reinforcement learning in recent years has been in application to so called \textit{episodic} environments. These are environments that exhibit a clear decomposable structure into time windows that are drawn from a stationary distribution. This assumption about an environment can be very useful when applicable as it allows us to consider the performance of our policy during a single episode as a sample from a random variable representing our full objective: 

\begin{equation} \label{episodicobjective}
\begin{split}
J_{\text{episodic}}(\pi) &= v_\pi(s) = \mathbb{E}_{\pi}  \bigg[ \sum_{k=0}^{H-1} \gamma^k R_{t+k} \bigg| S_t = s \bigg] \\ 
&= J_{\text{continuing}}(\pi) - \mathbb{E}_{\pi}  \bigg[ \sum_{k=H}^{\infty} \gamma^k R_{t+k} \bigg| S_t = s \bigg]
\end{split}
\end{equation} 

The key difference between equation \ref{infinitehorizonobjective} and equation \ref{episodicobjective} is that the latter only optimizes over a future horizon $H$ until the current episode terminates rather than until the end of the agent's lifetime. Unfortunately, approaches developed for this setting experience biased optimization when applied to continual episodic reinforcement learning settings where episodes are drawn from a non-stationary distribution that changes over time. In these settings, because the stationary episodic structure assumption is no longer valid, we must consider our objective in the continuing setting from equation \ref{infinitehorizonobjective}. As such, it is clear that the standard episodic objective is biased to only optimize on the current episode distribution while disregarding the likely far more important expected future episode distribution over the agent's lifetime. As a result, it becomes clear that blindly applying episodic reinforcement learning approaches in this way to non-stationary settings leads to biased optimization, which is likely to cause catastrophic forgetting effects, for example, when we sample repeatedly from each task before changing task distributions. 

This kind of situation appears naturally in many real-world applications of continual RL. For example, let us consider an agent learning in an environment whose dynamics can be naturally broken down into several modes of operation that repeat over the course of an hour, day, or week. One example would be a taxi driving agent that experiences very different traffic and demand patterns depending on the time of day (i.e. morning rush hour, lunch-time, evening rush hour, etc.). If updates to our agent's policy are greedy based on only the current mode of operation, optimization will become quite difficult as many learning steps may be taken for a single mode without regard for how an agent performs on the other modes of operation. Because these modes are temporally correlated, this greedy optimization will directly encourage catastrophic forgetting of other modes during the learning of the current mode. On the other hand, an agent that correctly identifies that each mode of operation always constitutes a constant proportion of its expected future distribution of modes can perform balanced updates that do not overvalue the current mode of operation. Indeed, these balanced updates are a natural consequence of optimizing for the long-term objective in equation \ref{infinitehorizonobjective} rather than myopic short term optimization as in equation \ref{episodicobjective}.

\paragraph{Catastrophic Forgetting in Continual Supervised Learning}
In applications of supervised learning to non-stationary data distributions, we actually see a very similar phenomenon that leads to catastrophic forgetting. Gradient based algorithms like SGD and popular variants of SGD either implicitly or explicitly assume that they are optimizing over a stationary i.i.d. distribution of data. When we optimize over a non-stationary distribution of data, we can see SGD as related to the deterministic policy gradient theorem with a differentiable reward function (i.e. the loss function). However, SGD simply optimizes on the current distribution of experiences without regard for the expected future distribution of experiences. As a result, when standard approaches to supervised learning are applied to non-stationary setting such as the popular \textit{locally i.i.d.} setting \citep{kirkpatrick2017overcoming,GEM,MER,AGEM}, it is easy to see how the optimization is biased to only update for the current distribution while potentially catastrophically forgetting its capabilities on old data distributions that may once again be relevant in the future. See Appendix \ref{app:supervisedlearning} for a more detailed description of how to view continual supervised learning as a special case of RL in a continuing environment. 

\begin{figure}[h]
  \begin{center}
  \centerline{\includegraphics[width=1.0\linewidth]{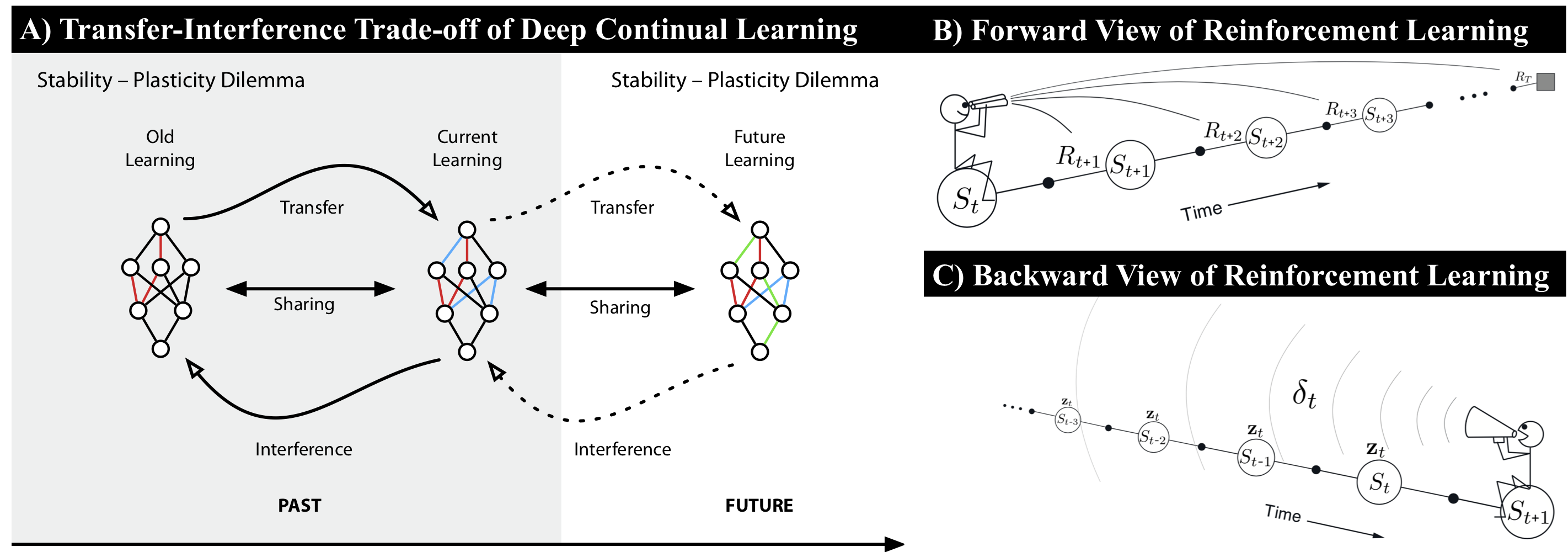}}
   \caption{\textbf{Reinforcement Learning and the Stability-Plasticity Dilemma}: A) Depicts the stability-plasticity dilemma and its relation to both weight sharing and transfer dynamics over time (from \citep{MER}). B) Depicts the forward view of RL where we evaluate the current state based on expected future rewards (from \citep{Sutton98}). C) Depicts the backward view of RL where we leverage recent states and rewards to correct our evaluations of past states (from \citep{Sutton98}).} %
   \label{fig:forwardbackward}
   \end{center}
   \vspace{-20pt}
\end{figure}

\paragraph{Studying Time Correlations in CL with RL}
As we just discussed, a key assumption in supervised learning is that data is independently and identically distributed (i.i.d) and drawn from a fixed distribution. However, in reinforcement learning, data points are not only time correlated, but also, do not come from a fixed distribution. Considering that data points are not i.i.d, studying the learning dynamics within RL facilitates a deeper understanding of long-term memory. In particular, a common issue in supervised CL is to determine how newly seen data might be related to previously seen data. However, it is counter intuitive to analyze this (as is typically done) without an explicit dependency on time. We should note that one could always formulate a supervised learning problem within the RL framework as highlighted by \citep{barto2004reinforcement}.

\paragraph{Understanding the Forward and Backward view of CL with RL}
RL as a paradigm offers algorithms which are forward focusing or backward focusing~\citep{Sutton98}. For each state visited, the forward view allows the agent to look forward in time to consider the future rewards and decide how best to combine them. Due to the unavailability of future states a more efficient incremental strategy makes use of backward-view computations. One could imagine that typical problems of balancing the stability-plasticity dilemma could potentially be more naturally understood with the forward and backward views readily available in RL. For example, in Figure \ref{fig:forwardbackward} we highlight how the ideas of forward and backward transfer and interference in continual learning \citep{MER} are naturally subsumed by RL's notion of a forward and backward view. 

\section{A Taxonomy of Continual RL Problem Formalisms} 
\label{sec:taxonomyformalism}
Due to the generality of the continual learning problem, formulations vary vastly in the literature as we also highlight in the former sections. In this section, we provide a taxonomy of the CRL problem formulations. Foremost, we consider the setting where all components of the problem can take a non-stationary functional form $f(i,t)$ as the most general continual reinforcement learning problem. Moreover, we detail some key additional assumptions on the functional form of non-stationarity that have been prominent in the literaure.

\subsection{General CRL Problem}
Formally, we understand a \texttt{General CRL Problem} $\mathcal{M}_\texttt{CRL}$ as:

\begin{mdframed}
\ddef{\texttt{General CRL Problem} $\mathcal{M}_\texttt{CRL}$}{Given a state space $\mathcal{S}$, action-space $\mathcal{A}$, an observation space $\mathcal{O}$, a reward function $r: \mathcal{S} \times \mathcal{A} \rightarrow \mathrm{R}$, a transition function $p: \mathcal{S} \times \mathcal{A} \rightarrow \mathcal{S}$, and an observation function $x: \mathcal{S} \rightarrow  \mathcal{O}$, the most general continual reinforcement learning problem can be expressed as:
\begin{align}
    \mathcal{M}_\texttt{CRL} \doteq \langle \mathcal{S}(t), \mathcal{A}(t), r(t), p(t) , x(t), \mathcal{O}(t)
    \rangle,
\end{align}
where each component of the problem formulation can be considered as a non-stationary function of form $f(i,t)$ where $i$ is the input specific to each component. 
\label{def:general-crl-problem}
}
\end{mdframed}

\subsection{Common Non-stationary Functional Forms}
Due to the very broad nature of the \texttt{General CRL Problem} in the definition above, %
concrete assumptions on the non-stationary functional form $f(i,t)$ are typical in the literature to help define structure in non-stationarity. This is indeed critical as an arbitrarily non-stationary environment gives no consistent signal to learn from, making learning impossible. As such, not making additional functional assumptions about the nature of non-stationarity can lead to a vacuous problem statement.
The most common types of assumptions about non-stationarity in the literature include Lipschitz continuity and piecewise non-stationarity.

\begin{mdframed}
\ddef{\texttt{Non-stationary Functions With Lipschitz Continuity} $f(i,t)$}{A function is Lipschitz non-stationary if across the input space $i \in \mathcal{I}$ non-stationary can be bounded by a time independent constant $C$:
\begin{align}
\forall i \in \mathcal{I}, \forall t,t' \in \mathbb{R} \quad |f(i,t) - f(i,t')| \leq C|t - t'|. 
\end{align}
}
This condition also ensures the function has bounded first derivatives in time $\partial f / \partial t$.
\end{mdframed}

\pagebreak
\begin{mdframed}
\ddef{\texttt{Piecewise Non-stationary Functions} $f(i,t)$}{A function is piecewise non-stationary if it can be seen as broken up into stationary functions $f_0(i),f_1(i),f_2(i),...$ of the input space $i \in \mathcal{I}$ over intervals of time dictated by $t_0,t_1,t_2,...$ such that:
\begin{align}
\forall i \in \mathcal{I} \quad f(i,t) = \begin{cases}
  f_0(i)  & 0 \leq t < t_0 \\
  f_1(i)  & t_0 \leq t < t_1 \\ 
  f_2(i)  & t_1 \leq t < t_2 \\ 
  ...
\end{cases}
\end{align}
}
\end{mdframed}

These are simply common examples of assumptions about the nature of non-stationarity. For example, another potentially interesting formulation is to assume arbitrary non-stationarity within a fixed variation budget as in \citep{mao2021near}.

\subsection{Key Properties of Non-stationarity: Scope and Drivers}
We would like to provide a taxonomy of formulations that includes prominent assumptions about non-stationarity that have been either explicitly or implicitly considered in the literature. Towards this end, we present a categorisation of non-stationarity along \underline{two primary dimensions}, namely the 
\textbf{scope} and \textbf{driver} of non-stationarity. 

\begin{mdframed}
\ddef{Scope of non-stationarity $\alpha$}{defines what elements of the agent-environment interaction process experience non-stationarity:
\begin{align}
\alpha \subseteq \{ \mathcal{S}, \mathcal{A}, r, p , x, \mathcal{O} \}, 
\end{align}
where $p \in \alpha \text{ if  }  \exists   t, t' \in \mathrm{R}, p(t) != p(t'), $ $r \in \alpha \text{ if  }  \exists   t, t' \in \mathrm{R}, r(t) != r(t') $, etc.
}
\end{mdframed}

\begin{mdframed}
\ddef{Driver of non-stationarity $\beta$}{defines the causal assumptions that can be made about the nature of the evolution of non-stationary environment dynamics: 
\begin{align}
    \beta \in \{ \texttt{stationary}, \texttt{passive}, \texttt{active} , \texttt{hybrid} \},
\end{align}
where \texttt{stationary} $\implies$  $\mathbb{E}[f(i,t)] = \mathbb{E}[f(i,t’)] \;\; \forall t \in \mathbb{R}, \forall t' > t, \;\; \forall i \in \mathcal{I}$, \\
\texttt{passive} $\implies$ if $\mathbb{E}[f(i,t)] \neq \mathbb{E}[f(i,t’)]$, then  $|\mathbb{E}[f(i,t)] - \mathbb{E}[f(i,t’)]| \indep a \;\; \forall a \in \mathcal{A}, \;\; \forall t \in \mathbb{R}, \forall t' > t, \;\; \forall i \in \mathcal{I}$, \\
\texttt{active} $\implies$ if $\mathbb{E}[f(i,t)] \neq \mathbb{E}[f(i,t’)]$, then  $|\mathbb{E}[f(i,t)] - \mathbb{E}[f(i,t’)]| \not \indep a \;\; \forall a \in \mathcal{A}, \;\; \forall t \in \mathbb{R}, \forall t' > t, \;\; \forall i \in \mathcal{I}$, and  \\
\texttt{hybrid} $\implies$ if $\mathbb{E}[f(i,t)] \neq \mathbb{E}[f(i,t’)]$, then  $|\mathbb{E}[f(i,t)] - \mathbb{E}[f(i,t’)]| \indep a \;\; \exists a \in \mathcal{A}, \;\; \exists t \in \mathbb{R}, \forall t' > t, \;\; \exists i \in \mathcal{I}$ and
$|\mathbb{E}[f(i,t)] - \mathbb{E}[f(i,t’)]| \not \indep a \;\; \exists a \in \mathcal{A}, \;\; \exists t \in \mathbb{R}, \forall t' > t, \;\; \exists i \in \mathcal{I}$.
}
\end{mdframed}

\subsection{Examples of CRL Formulations}

We now discuss how the proposed taxonomy can provide a lens to investigate existing formulations. Coupled with the consideration of the driver and scope of the non-stationarity, the \texttt{General CRL Problem} $\mathcal{M}_\texttt{CRL}$ can be cast as \textbf{a family of MDPs with non-stationary functional forms} (See Proposition \ref{prop:crl-as-nonstationaryMDP}). Moreover, such a non-stationarity MDP can itself be recast as a \textbf{partially-observable MDP} (See Proposition \ref{prop:nonstationaryMDP-POMDP-duality}) resulting in the following CRL problem formulations:

\begin{mdframed}
\begin{proposition}[Non-stationary MDPs as \texttt{CRL Problems}]
\label{prop:crl-as-nonstationaryMDP}
A \textbf{non-stationary MDP} is a special type of CRL problem where $\alpha \subseteq \{ \mathcal{S}, \mathcal{A}, r, p \}$, the observation function is an appropriate identity matrix $x = \mathbb{I}$, and the observation space is the state space $\mathcal{O} = \mathcal{S}$.
\end{proposition}
\end{mdframed}

In this partially observable view, an arbitrarily non-stationary MDP is still unwieldy to solve as it corresponds to an infinite order history dependence. However, in many cases the partial observability view may be preferable for theoretical analysis and crafting more aggressive approaches. This is because this formulation allows for more aggressive solutions as epistemic uncertainty about the future reduces in comparison to assumptions such as piecewise and Lipschitz non-stationarity which always take a local view and never arrive at certainty about the nature of the global problem. It is also amenable to theoretical analysis as the greater system is considered stationary and has well defined long-term behavior. 

\begin{mdframed}
\begin{proposition}[Non-stationary MDP and POMDP Duality]
\label{prop:nonstationaryMDP-POMDP-duality}
    Any non-stationary MDP $\mathcal{M}$ with $\alpha \subseteq \{ p, r\}$ can dually be viewed as an equivalent POMDP $\hat{\mathcal{M}}$. This is because the potentially non-stationary transitions $p(s’|s,a,t)$ and rewards $r(s,a,t)$ of $\mathcal{M}$ can appear stationary with a simple change of variables to create $\hat{\mathcal{M}}$ with observation $\hat{o}=s$ and a full state also based on the unobserved time dependent variable $\hat{s}=[\hat{o},t]$ so that transitions are $p(\hat{s}'|\hat{s},a)$ and rewards are $r(\hat{s},a)$. Furthermore, any POMDP can be seen as a non-stationary MDP because the combination of the observation and time variables always constitutes a uniquely identifying state representation.
\end{proposition}
\end{mdframed}

\textbf{Multi-agent RL Example:} As an example of an MDP that can be either viewed as non-stationary or partially observable depending on our perspective of the problem, we will briefly consider a multi-agent environment where each agent is constantly learning. Multi-agent environments are often characterized as a \textit{Markov game} \citep{littman94markov} in which a set of agents all interact in the environment and jointly impact the reward and transition dynamics. Formally, the reward dynamics $r(s,a^i,\bm{a}^{-i})$ and transition dynamics $p(s'|s,a^i,\bm{a}^{-i})$  depend on the global state of the environment $s$, $a^i$ which denotes the action of some agent of focus, and $\bm{a}^{-i}$ denoting the vector of actions for all other agents in the environment. If the policies of the other agents remain constant, we can view this formulation as equivalent to a stationary single agent learning problem such that $r(s,a^i)=\sum_{\bm{a}^{-i}} \pi(\bm{a}^{-i}|s) r(s,a^i,\bm{a}^{-i})$ and $p(s'|s,a^i)=\sum_{\bm{a}^{-i}} \pi(\bm{a}^{-i}|s) p(s'|s,a^i,\bm{a}^{-i})$ without at all considering the role of the other agents in the environment. However, if their policies change over time, the rewards $r(s,a^i)$ and transitions $p(s'|s,a^i)$ will become non-stationary from the perspective of the focal agent. This apparent non-stationarity from the perspective of agent $i$ was recently recast as partial observability using the formalism of an \textit{active Markov game} \citep{further}:

\pagebreak
\begin{mdframed}
\begin{proposition}[Active Markov Games as \texttt{CRL Problems}] \label{prop:activemarkovgame}
{Active Markov games define a set of problems that can be viewed as partially observable or time-dependent because of the dependence of $p$ and $r$ on the actions of other agents in the environment. The actions of other agents are not observed at the same time as the state, nor are their time-dependent parameters, nor their update functions.}
\end{proposition}
\end{mdframed}

As discussed in \citep{further}, finding the optimal policy in this environment can be viewed as finding the optimal \textbf{stationary periodic distribution}. Moreover, if each agent finds its own optimal non-stationary policy, the result is called an \textbf{active equilibrium}. If each agent finds its optimal stationary policy, the result is called a \textbf{Nash equilibrium}.

\subsection{A Unified View}

A \texttt{General CRL Problem} $\mathcal{M}_\texttt{CRL}$ broadly captures existing problem formulations in the literature. The two primary dimensions, \textbf{scope} and \textbf{driver} of non-stationarity, provide a taxonomy that can characterize different CRL problem formalims. This view results in CRL as a strict generalization of existing settings and therefore offers a unified formulation.

\begin{mdframed}
\begin{proposition}[CRL as Strict Generalization]
\label{prof:crl-as-strict-gen}
    The \texttt{CRL Problem} as stated in Definition 1 %
    is a strict generalization of existing categories of problem settings.\footnote{This view allows us to define categories of settings and not necessarily concrete setting descriptions.}
    \begin{enumerate}
        \item Multi-task MDPs are special CRL problems where $\beta \in \{ \texttt{stationary} \}$.
        \item HiP-MDPs \citep{doshi2013hidden} are special CRL problems where $\alpha \subseteq \{ p, r \}$ and $\beta \in \{ \texttt{stationary} \}$.
        \item HM-MDPs \citep{hmMDP} and DP-MDPs \citep{xie2020deep} are special CRL problems where $\alpha \subseteq \{ p, r \}$ and $\beta \in \{ \texttt{stationary}, \texttt{passive} \}$.
        \item MOMDPs \citep{MOMDP} and active Markov games \citep{further} are both special CRL problems where $\alpha \subseteq \{ p, r \}$ and \\$\beta \in \{ \texttt{stationary}, \texttt{passive}, \texttt{active}, \texttt{hybrid} \}$.
    \end{enumerate}
\end{proposition}
\end{mdframed}

\begin{figure}[h]
  \begin{center}
  \centerline{\includegraphics[width=1.0\linewidth]{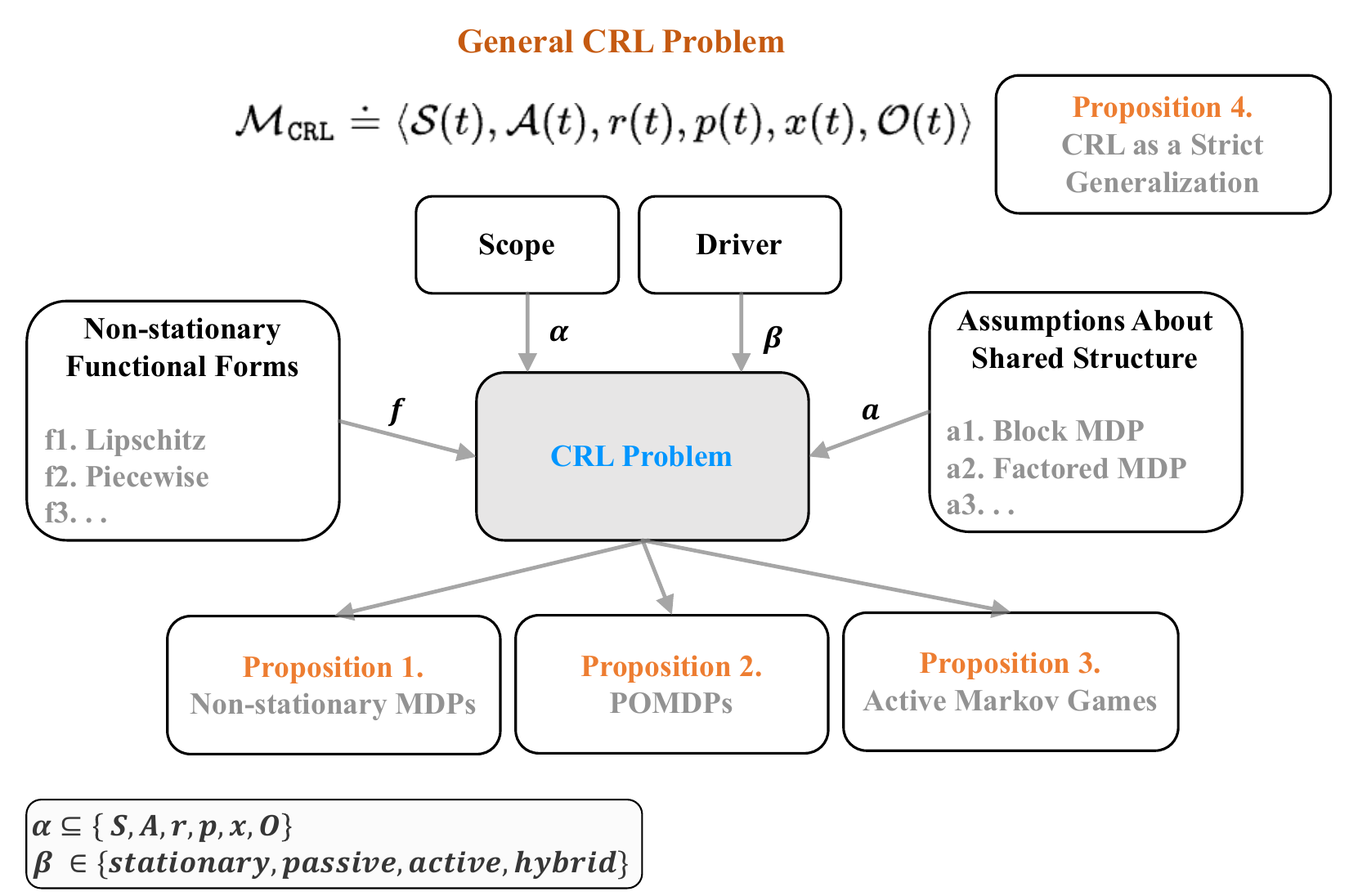}}
   \caption{\textbf{Taxonomy of Continual RL Formalisms}: Problem formulations in continual RL can be categorized along two primary dimensions: 1) the \textbf{scope} of the non-stationarity $\alpha$ and 2) the \textbf{driver} of non-stationarity $\beta$. Coupled with the scope and the driver of the non-stationarity, assumptions about the \textbf{non-stationary functional forms} ($f$) and \textbf{shared structure} ($a$) can result in different CRL formulations (Propositions \ref{prop:crl-as-nonstationaryMDP}, \ref{prop:nonstationaryMDP-POMDP-duality}, and \ref{prop:activemarkovgame}). This view offers a unified perspective resulting in continual reinforcement learning as a strict generalization of most of the existing formulations in the literature (Proposition ~\ref{prof:crl-as-strict-gen}). }
   \label{fig:taxonomy_formalisms}
   \end{center}
   \vspace{-30pt}
\end{figure}

\subsection{Assumptions About Shared Structure}

Next, it is common to assume shared structure across different components of the \texttt{CRL Problem} $\mathcal{M}_\texttt{CRL}$. This is because generic MDPs with no assumed structure have a minimax $T$-step regret bound of $\Omega(\sqrt{|\mathcal{S}||\mathcal{A}|T})$ and sample complexity bound of $\Omega(|\mathcal{S}||\mathcal{A}|)$ \citep{jaksch2010near,episodicbound}, which implies that no proveable generalization to novel state and action pairs is possible in this setting. 

Luckily, many real world applications also possess shared structure either in the observation space, state space, transition dynamics or reward dynamics. For instance, objects or abstract concepts emerge from shared structure across different regions of the world we live in. Similarly, laws of physics governing motion, force, or how objects interact are shared across different tasks and applications. We now detail a few popular assumptions that have emerged in the literature. This includes Block MDPs \citep{du2019provably}:

\begin{mdframed}
\ddef{Block MDP Assumption}{Each observation $o$ uniquely determines its generating state $s$. That is, the observation space $\mathcal{O}$ can be partitioned into disjoint blocks $\mathcal{O}_s$, each containing the support of the conditional distribution $x(\cdot|s)$. 
} \end{mdframed}

Moreover, in many real-world settings a slowly changing i.e. Lipschitz assumption may not be practical as changes may be large. However, it may still be the case that changes are limited to only a small subset of the causal variables of the underlying MDP \citep{gumbsch2021sparsely}. These cases can be formalized through the Factored MDP assumption~\citep{kearns1999efficient,BOUTILIER200049}: 

\pagebreak
\begin{mdframed}
\ddef{Factored MDP Assumption}{Each state is composed of $n$ factors $s := s^1,s^2,...,s^n \in \mathcal{S} \subseteq \mathcal{S}^1 \times \mathcal{S}^2 \times ... \times \mathcal{S}^n$ and each action can be composed of $m$ components $a := a^1,a^2,...,a^m \in \mathcal{A} \subseteq \mathcal{A}^1 \times \mathcal{A}^2 \times ... \times \mathcal{A}^m$ with sparse causal interactions such that $p(s'|s,a) = \prod_{i=1}^n p_i(s^{i'}|\text{Par}_i(s,a))$ where $\text{Par}_i$ sub-selects from the set of state and action components $\{1,...,n+m\}$ based on the causal Bayesian network structure. The reward function may also be factored such that $r(s,a) := (1 / \ell) \sum_{j=1}^{\ell} r_j(\text{Par}_j(s,a))$.
}
\end{mdframed}

Moreover, there are a number of other assumptions about share structure that have proven useful in the literature. For example, POMDPs can be restricted to those displaying linear structure as in \citet{cai2022sample}. Additionally, assumptions about low-rank structure enable a basis for provable generalization while allowing for more complexity than simpler Block MDP models \citep{agarwal2020flambe,uehara2021representation,zhang2021provably}. Similarly, recent work has achieved proveable generalization for problems with low Bellman Eluder dimensions \citep{jin2021bellman} and low dimensional underlying causal structure \citep{huang2021adarl}. It is also indeed possible to combine the Block MDP and Factored MDP assumptions into a single common formulation \citep{misra2021provable} to relax Block MDPs to handle non-stationary environments \citep{katt2019bayesian,sodhani2022block}. Additionally, Block MDPs have been formalized in settings with multiple domains \citep{han2021learning}. Furthermore, a number of settings exist that refine Factored MDPs even further in order to provide additional opportunities for generalization including relational MDPs \citep{guestrin2003generalizing}, first-order MDPs \citep{boutilier2001symbolic}, object-oriented MDPs \citep{diuk2008object}, and MDPs described by linear temporal logic \citep{vaezipoor2021ltl2action,jiang2021temporal}.

\subsection{Literature Review: The Scope of Non-stationarity $\alpha$} 

We now review the literature through the lens of the scope of the non-stationarity. Revisiting the definition of a MDP as a tuple $M = \langle \mathcal{S}, \mathcal{A}, r, p, \gamma \rangle$, it is quite common to assume that there is non-stationarity in either rewards or transition dynamics i.e. $\alpha \subseteq \{ p, r\}$. Meanwhile, it is also possible that the observation function $x$ or action space  $\mathcal{A}$ is non-stationary as highlighted in Figure \ref{agent-env-interface} i.e. $\alpha \subseteq \{ \mathcal{S}, \mathcal{A}, r, p, x, \mathcal{O}\}$. Very few researchers have actually explored this setting in a fully general manner with every component being potentially non-stationary to date (i.e. the scope). As we discussed earlier, this level of non-stationarity can be quite problematic for standard policy optimization approaches and lead to catastrophic optimization failures. As a result, we will begin by highlighting some prominent formulations in the literature with a smaller scope of non-stationarity.

Non-stationarity in the reward function alone can be modelled in the problem formulation through many different approaches. For instance, \textit{goal directed} learning without a task specific reward (see Sec.~\ref{sec:goalfocused}) could be interpreted as containing non-stationarity only in the reward function. These approaches allow the agent to learn about other stationary elements of the environment while being able to adapt as the reward function changes with changing goals. In fact, a family of related methods consider the \textit{unsupervised RL} setup (see Sec.~\ref{sec:goalfocused}) with the aim of learning task-agnostic behaviors, wherein the agent is trained with no explicit rewards, but later evaluated on specific tasks. The shift from no reward to a task-specific reward distribution could be interpreted as a source of non-stationarity that the agent is explicitly trained to prepare for by learning as much as it can about the common stationary elements of the environment. 

It is also somewhat common to consider environments where only the transition dynamics change and the reward function remains constant. Moreover, the most common case including two elements of non-stationarity is to assume that both the transition dynamics and reward function may change as is common in the meta-learning literature (see Sec.~\ref{sec:learningtoadapt}). This setting itself can be considered quite challenging and is probably the most ambitious formulation that is commonly attempted in the literature. It should be noted that most of these papers do not explicitly consider the continual RL setting, but the concepts could be extended to study this more general form of agent deployment.

While less common, work on non-stationary RL has also considered the case of a changing action space \citep{chandak2019learning,langlois2021rl,jain2021know,trabucco2022anymorph} and observation function \citep{zhang2020invariant,trabucco2022anymorph}. These sources of non-stationarity may be very important for particular practical applications of continual RL. However, these remain an under explored problem settings in comparison to those focused on transition and reward dynamics. Moreover, to date, we are not aware of any approach that allows for every element of the problem formulation to be non-stationary. It remains an open question to what degree learning is still possible when we allow for non-stationarity in each component of the environment and it is likely that additional assumptions about the nature of the problem will be necessary to develop agents that perform well in this very ambitious setting.

\subsection{Literature Review: The Drivers of Non-stationarity}  

An important point of distinction between different formalism for continual RL is what assumption is made with regard to the agent's own causal role in influencing the non-stationary evolution of the environment. First, we should note that it is possible that the non-stationarity of the environment is drawn from a stationary distribution that the agent cannot influence. Additionally, non-stationary settings that are influenced by the agent's behavior can be said to be active while settings where the non-stationarity is independent of the agent's behavior can be considered passive. It is also possible that the environment may be non-stationary both based on the agent's behavior and based on causal factors beyond the agent's control. 

\subsubsection{Stationary Task Distributions: Multi-Task Learning} 

Perhaps the most common assumption in the literature is that while there may be multiple tasks $z$, which each have their own associated MDP $M^{(z)}$, these tasks are sampled from a unknown but {\em fixed} distribution $p(z)$. This assumption is common for learning in settings such as multi-task learning \citep{Caruana97, wilson2007multi, ammar2014online}. This setting can be far more difficult than single task learning as the environments may require diverse and possibly conflicting behaviors to achieve success. This setting can also be viewed with the lens of multi-objective optimization where potentially conflicting requirements make it important to consider concepts like \textit{Pareto Optimality} \citep{gabor1998multi} for cases where it is not possible to learn an optimal policy for each task with the same shared set of parameters. Additionally, when gradient based learning and function approximation are applied to multiple diverse tasks simultaneously, this is known to often result in interfering gradients~\citep{french1991using,Caruana97} across tasks. As noted by \citep{gradientsurgery}, conflicting gradients can be particularly damaging to the learning process when tasks have differing gradient magnitudes and when the loss landscape exhibits high curvature. 

A large body of work is focused on online multi-task learning where tasks are faced in a sequential fashion. A class of algorithms for multi-task RL in this setting use non parametric Bayesian models facilitating knowledge sharing across tasks. One approach~\citep{wilson2007multi, ammar2014online} is to model the distribution over tasks and use this distribution as a prior when a new task is seen. In the real world, data is often collected from experiences in many different environments and it is desirable to share knowledge learned from previous experiences \citep{li2009multi}.

Recently popular approaches for "meta-learning" \citep{schmidhuber1987evolutionary, bengio1992optimization} have come to leverage protocols for what is called \textit{meta-training} and \textit{meta-testing} that also align with this assumption of a stationary task distribution $p(z)$ \citep{wang2016learning,duan2016rl,MAML,snail,Reptile}. See \citep{vilalta2002perspective,hospedales2020meta} for an in-depth survey of meta-learning. In \textit{meta-training} these meta-learning approaches learn a policy that can quickly adapt to tasks in the distribution $p(z)$. While it is generally assumed that during \textit{meta-testing} different tasks are used for learning than those learned in \textit{meta-training}, we only expect this generalization to be possible to the extent that there are shared commonalities between the \textit{meta-testing} tasks and the \textit{meta-training} distribution $p(z)$. In fact, there is often nothing done to explicitly prepare for out of distribution generalization or distributional shifts. See Sec.~\ref{sec:learningtoadapt} for a more in depth discussion of this problem setup.

\subsubsection{Passive Non-stationarity}

In passive non-stationary environments, we assume that the non-stationary behavior (i.e. the evolution of tasks) does not depend on the behavior of the agent itself when interacting with the environment. This allows us to model the evolution of tasks using the stochastic function $p(z'|z)$ as in \citep{hmMDP} without having to consider the effects of our own changing policy on this distribution. While this setting is less human realistic in some sense, it is quite practical and describes the setting of the clear majority of experiments in the continual RL literature. An extension of this idea is to consider task transitions that happen at irregular multi-task intervals in the Semi-Markov Decision Process \citep{sutton1999between} setting as in \citep{hadoux2014solving}. For example, this setting can model tasks that transition after the termination of each episode. Indeed, much of the work on learning to infer latent contexts (see ~\ref{sec:contextdetection}) and learning to adapt (see~\ref{sec:learningtoadapt}) can be seen as part of this category of approaches that assume a passive source of non-stationarity.  

One kind of popular assumption that makes solving passive non-stationary MDPs tenable is local consistency. Even mild assumptions such as restricting the change in an MDP to be slow or Lipschitz can provide a basis for robust worst case optimization \citep{lecarpentier2019non,li2021dealing}. Another assumption that is common is that of piecewise stationarity \citep{nonstationaryrlenvs} or local stationarity with change points. Generally approaches in these settings mentioned consider the problem of learning about the dynamics of $p(z'|z)$ to be too challenging and instead adopt a purely reactive philosophy with respect to environment changes. Alternatively, to account for both slow and abrupt changes in transition and reward dynamics, one approach is to quantify variation in the reward function and transition kernel over time in terms of their respective variation budgets $\Delta_r$, $\Delta_p$. In this framework, it then possible to consider assumptions around variation budgets in tabular~\citep{cheung2020reinforcement, domingues2020kernel} and linear function~\citep{touati2020efficient} MDPs to improve learning in the face of non-stationarity.

\subsubsection{Active Non-stationarity}

In active non-stationary environments, we consider that our behavior may have an impact on the nature of the non-stationarity in the environment itself. This concept is foundational to work on intrinsic curiosity~\citep{schmidhuber1991curious,chentanez2005intrinsically, singh2009rewards, barto2013intrinsic, achiam2017surprise} and learning an agent's own curriculum~\citep{powerplay,justesen2018automated, srinivasan2019automated}. See \citep{curriculum} for a survey of automated curriculum learning approaches for deep RL. For example, agents have in some settings learned to generate their own imagined environments or scenarios for future play \citep{POET,raileanu2021automatic,chen2021variational,lee2021improving}, their own goals~\citep{GOAL1,GOAL2,sharma2021autonomous}, relevant information from their past learning \citep{klink2021boosted,jiang2021replay}, and their own opponents~\citep{sukhbaatar2017intrinsic}. Agents in ambitious open-ended learning settings, where both the agent and the designer do not know the task or the domain ahead of time, could also experience active sources of non-stationarity. For example, these settings may experience a changing state and action space over the course of an agent's lifetime while continuously generating creative behaviors~\citep{doncieux2020dream}. In active non-stationary environments, we can now assume that environment dynamics may vary in a Markovian way following the function $p(z'|s,a,z)$ as in~\citep{MOMDP}. See Sec.~\ref{sec:learningtoexplore} for instances of approaches attempting to solve such a formulation.

\subsubsection{Active and Passive Non-stationarity}

It is also possible to consider settings where the non-stationarity can be controlled in some ways by changing the agent's behavior, but also influenced by causal mechanisms beyond the agent's control. For example, in multi-agent RL an agent may be able to influence the learned behavior of another agent by changing its own behavior and play an active role in shaping the non-stationarity of the RL problem from its perspective~\citep{lola,loladice,metamapg,further}. However, certain aspects of the non-stationary evolution of the environment based on changes in multi-agent behavior is likely beyond the control of each individual agent, which is an important consideration when designing continual models. This hybrid non-stationary setting, while less common in the literature to date, seems like the most representative of many real-world applications. As such, designing agents that can function in this kind of non-stationary environment represents an emerging frontier for increasingly ambitious continual RL research that can tackle real-world problems. 

\section{A Taxonomy of Continual RL Approaches}
\label{sec:taxonomyapproaches}

\begin{figure}[h]
  \begin{center}  \centerline{\includegraphics[width=1.0\linewidth]{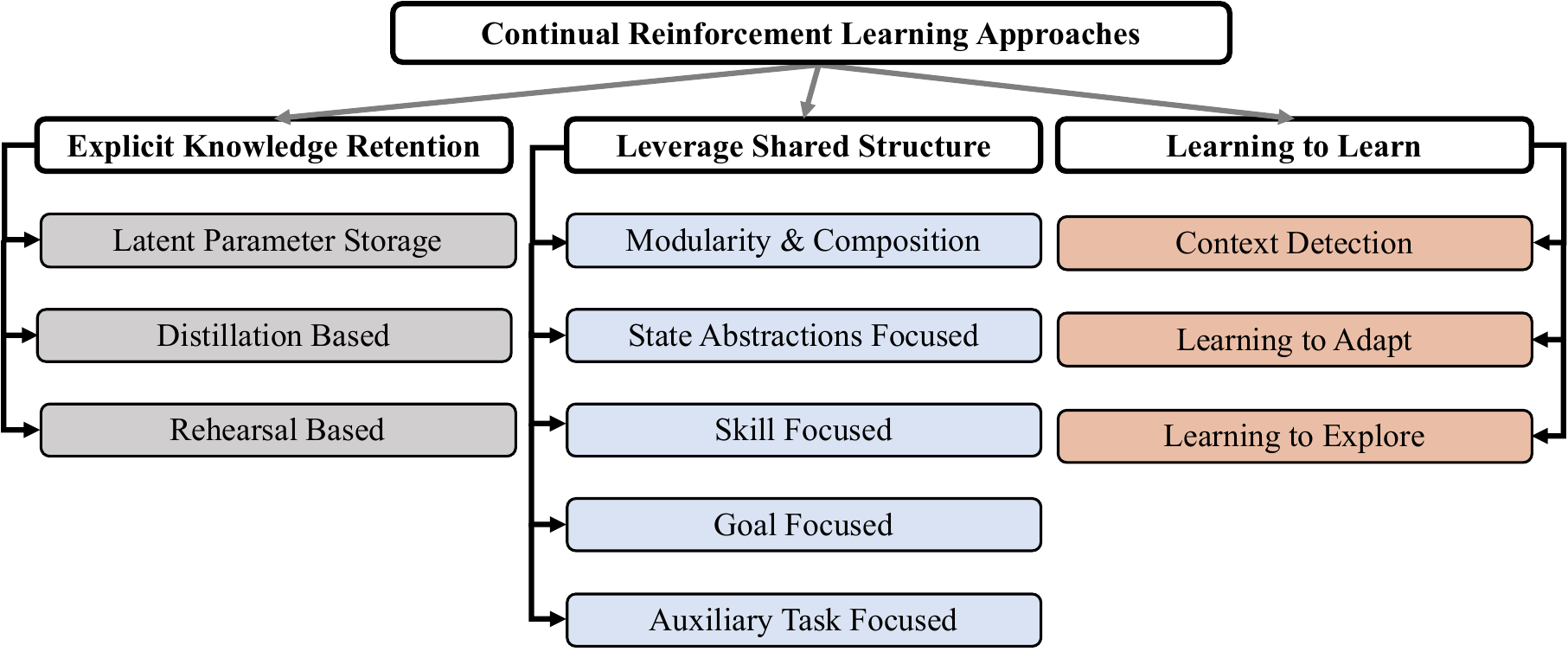}}
   \caption{\textbf{Taxonomy of Continual RL Approaches}: A diagram illustrating different clusters of approaches for continual RL, highlighting prominent threads of research within each family. Though these categories are not mutually exclusive, we examine each separately for the purpose of this paper.}
   \label{fig:taxonomy_approaches}
   \end{center}
   \vspace{-20pt}
\end{figure}

In this section, we discuss a taxonomy of approaches for continual RL as highlighted in Figure \ref{fig:taxonomy_approaches}. We will describe three high-level clusters: those focused on explicit knowledge retention, those focused on leveraging shared structure, and those focused on learning to learn. 

\begin{figure}[h]
  \begin{center}  \centerline{\includegraphics[width=1.0\linewidth]{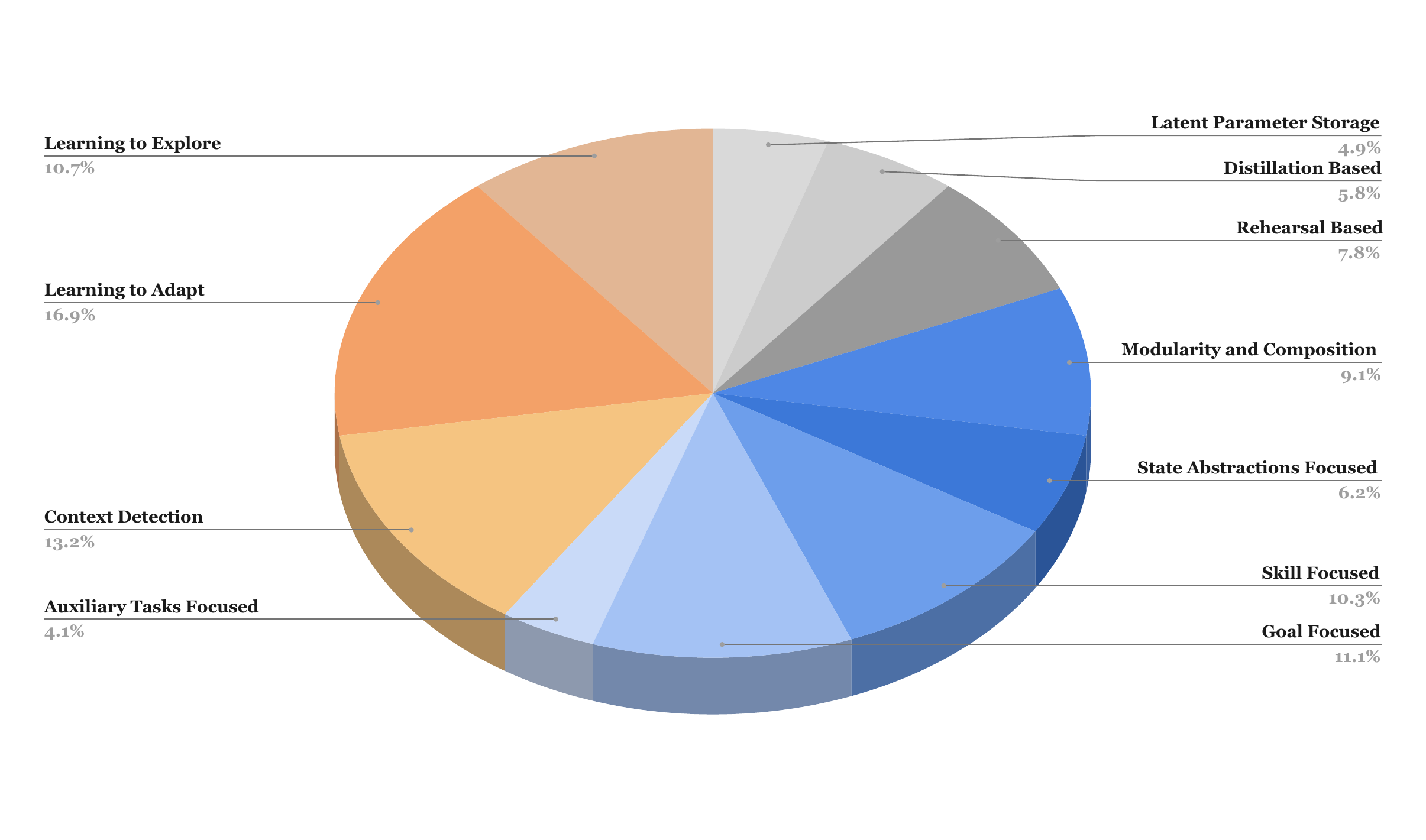}}
   \caption{\textbf{Popularity of Subcategories of Continual RL Approaches}. This chart tallies the citations in our paper by subcategory. This is not meant to be exhaustive or exact, but rather provide a high-level idea of the relative popularity of these topics in the continual RL literature to date.}
   \label{fig:taxonomy_approaches_visual}
   \end{center}
   \vspace{-20pt}
\end{figure}

Figure \ref{fig:taxonomy_approaches_visual} depicts the distribution of subcategories of approaches for continual RL. The taxonomy of approaches broadly consists of three key families: explicit knowledge retention (18.5\%), leveraging shared structure (40.8\%), and learning to learn (40.7\%). Across subcategories, we note that, unsurprisingly, learning to adapt is predominant. On the other hand, techniques based on latent parameter storage, distillation, state abstraction, and auxiliary tasks constitute only a small fraction of the approaches covered.

\subsection{Explicit Knowledge Retention}

Due to concerns about agents experiencing catastrophic forgetting when they continually learn as detailed in Sec. \ref{sec:posingrl}, a number of approaches have been proposed for \textit{stabilizing} learning based on a belief that straightforward optimization displays too much \textit{plasticity} in this setting. 

\subsubsection{Parameter Storage Based}

The most obvious way to prevent catastrophic forgetting across tasks in continual learning is to save an independent model for each task. However, this is sub-optimal for a number of reasons. First of all, it requires a methodology for detecting the current task. It also requires significant storage, and lastly, it limits any ability to leverage relevant knowledge learned across tasks. One alternative way to leverage knowledge is through the use of shared latent components. \citet{ammar2014online} accomplish this by using a shared latent basis that captures reusable components of the learned policies. Leveraging the assumption of shared common structure between tasks, \citet{borsa2016learning} build on the work of \citet{calandriello2014sparse} and explicitly model a shared abstraction of the state-action space. %

Another approach for leveraging knowledge from previous tasks is to provide the representations of networks trained on previous tasks as inputs for subsequent tasks~\citep{PNN}. While this strategy directly avoids the problem of catastrophic forgetting, it exacerbates the curse of input dimensionality and storage requirements. As the number of tasks seen grows, there is a larger space of past representations to sift through. As such, there is still some degradation in the efficacy of transfer in practice even if catastrophic forgetting is circumvented. One way to overcome these issues is by considering a single shared representation. For example, \citet{maurerBenefitMultitaskRepresentation} extracted features for multiple tasks in a single low-dimensional shared representation. \citet{D'Eramo2020Sharing,shi2021meta} further highlight the benefits of learning a shared representation, as error propagation in approximate value iteration and policy iteration improves when learning multiple tasks jointly. 

Alternatively, a popular approach is to store a prior about the extent of past usage of each parameter during learning in order to preserve important old knowledge~\citep{kirkpatrick2017overcoming,pcgrad,liu2021conflict}. This kind of approach generally decreases \textit{plasticity} in areas of the network that were heavily used for past tasks, which can effectively prevent forgetting. Unfortunately, this \textit{stability} may also limit the potential for backward transfer in the process. A related approach achieves a similar goal by leveraging the concept of superposition, where context information is stored for each task, so that the weights can be explicitly decomposed into orthogonal sub-networks~\citep{cheung2019superposition,wortsman2020supermasks}. Minimizing representational overlap has the positive effect of minimizing forgetting, but it may also minimize the potential for transfer similarly to the single task learning case. As a result, care must be taken while leveraging this kind of approach in a continual learning context

\subsubsection{Distillation Based}

Another common way to encourage retention of knowledge from past tasks is by leveraging knowledge distillation~\citep{Caruana06,hinton2015distilling} from past tasks when learning a new task to prevent catastrophic forgetting as in~\citep{rusu2015policy,LwF,riemer16,impala,schwarz2018progress,berseth2018progressive,kaplanis2019policy,traore2019discorl,tirumala2019exploiting,igl2021transient,lan2022memory,zhang2022catastrophic}. Knowledge distillation refers to the process of using one neural network as a target or soft target for another. Distillation can be used to augment experiences for training a network by providing a new auxiliary target for the network being trained to match. 

In the context of RL, this target could refer to either a policy or value function. In a continual learning context, distillation is a popular strategy for implementing conservative updates so that the agent's learning has \textit{stability} in preserving important knowledge of past tasks whenever possible. An additional benefit of this distillation approach is that it ultimately learns a separate model for each task, which could help address issues of Pareto optimality when an agent must learn conflicting tasks. On the other hand, this results in the need for a knowledge compression strategy for distillation approaches to scale to true many task learning settings.

\subsubsection{Rehearsal Based} 

Another popular strategy for reinforcing the importance of experiences from the past distribution during continual RL is leveraging experience replay~\citep{Lin92}. As explained in~\citep{pan2018organizing}, experience replay is closely related to model-based RL approaches like Dyna~\citep{Dyna}, where a buffer is used to generate realistic samples from the past distribution of experiences. Replay approaches can thus help correct for the short-term bias in our objective function to the extent that the past is a good proxy for the future. As a result, replay has become a very successful approach for tackling continual RL~\citep{isele2018selective,MER,rolnick2019experience,oh2021model,henning2021posterior,liu2021regret,queeney2021generalized,chandak2021universal,lampinen2021towards,venuto2021policy,liotet2022lifelong}. However, replay may result in significant storage requirements as we scale to progressively more complex continual learning settings. This has led some to explore replacing replay buffers with \textit{pseudo-rehearsals} sampled from a generative model~\citep{Robins95,atkinson2018pseudo,daniels2022model}.

Another promising strategy for improving the storage efficiency of replay buffers is to directly learn to compress experiences during learning. This makes the buffer much more efficient, though now only consisting of approximate recollections~\citep{riemer2017scalable,caccia2019online}. While they have been quite successful relative to other approaches to date, replay based strategies generally struggle to effectively leverage past data when the behavior of the current policy is significantly different, as off-policy learning tends to be difficult in this setting (see~\citep{levine2020offline}). Additionally, it may not be necessary to perform replay to combat time correlation during learning \citep{kaplanis2018continual}. 

\subsection{Leveraging Shared Structure}
\label{sec:sharedstructure}
Continually learning agents must learn to solve problems so that they are able to find structure in the world that will benefit them later \citep{thrun1996discovering}. To achieve this, continual learning agents should reuse aspects of the solutions to previously solved subproblems through function composition \citep{griffiths2019doing} by abstracting relevant meaningful information in the form of abstract concepts \citep{parr1998reinforcement} or skills \citep{thrun1995finding}. Such an ability is seamless in humans; when performing a complex task, it is natural for us to automatically break the task into smaller subtasks and be able to plan, learn, and reason with knowledge represented across multiple timescales. Enabling similar abilities in continual RL agents will be crucial for representing knowledge in a way that promotes retention and transfer across the lifetime of an agent. In this section, we will discuss approaches in the literature for leveraging this kind of shared structure. 

\subsubsection{Modularity and Composition Focused}
\label{sec:modularityfocused}
A key idea explored in the literature is to explicitly formulate a family of tasks as compositionally-structured. The hope is that approaches in this setting can tackle the problem of how to build machines that are capable of compositional generalization. Compositional generalization can be understood as leveraging prior experience to solve compositional perturbations of prior problems or even more complex problems than the agent has seen to this point. In this spirit, early work by \citep{singh1992transfer} defined a class of composite tasks called sequential decision tasks, which are expressed as a temporal concatenation of simpler tasks. \citet{doya2002multiple} decompose a non-stationary task into multiple domains in space and time to allow for predictable environment dynamics associated with each of the agent's learned modules. Their work further highlights and corroborates findings from the neuroscience literature (see Sec~\ref{sec:neuro}). 

More recent work \citep{devin2017learning, frans2017meta, fernando2017pathnet, RoutingNets, meyerson2017beyond, kirsch2018modular, ramachandran2018diversity, chang2018automatically, liang2018evolutionary, alet2018modular, cases2019recursive, yang2020multi,lee2021sharing,tseng2021toward,mendez2022reuse,mendez2022modular,gaya2022building} has focused on training neural network modules which can be composed for a family of related tasks, leveraging a combination of modules specialized to each task. This work is deeply related to RL methods for neural architecture search \citep{NAS} and their one shot \citep{brock2017smash}, multi-task learning \citep{pasunuru2019continual}, and continual learning \citep{xu2018reinforced,pasunuru2019continual} variants. 

While these models have increased power of composition and increased leverage in avoiding negative transfer by dividing information between modules, there are a number of challenges that make this idea difficult to implement in a continual learning setting. For example, one such problem is the "chicken and egg" problem of learning modules to combine and learning how to combine them. Here the issue becomes that even if the environment is stationary, the process of both learning modules and learning to combine them become non-stationary as modules are continually updated and leveraged in new combinations. See \citep{rosenbaum2019routing} for a detailed survey of the motivations and challenges of these modular and compositional approaches.

\subsubsection{State Abstractions Focused}
\label{sec:stateabstractionsfocused}
State abstraction (or aggregation) is central to the idea of capturing common structure within various tasks and potentially facilitating positive forward transfer across related tasks. \citet{li2006towards} provide a unified view on the theory of state abstraction and consider a variety of metrics that can be used to help develop these abstractions. Given MDP $M$, with its abstracted version $\hat{M}$, the abstraction function $\phi: S \rightarrow \hat{S}$ maps states in the ground MDP to states in the abstract MDP. A useful abstraction preserves information that is crucial for the original MDP, or a family of MDPs. 

One such abstraction is state abstractions based on the PAC framework. A PAC state abstraction is defined such that it achieves correct clustering with high probability with respect to a distribution over learning problems~\citep{abel2018salrl}. In the context of lifelong learning, PAC state abstractions are guaranteed to hold with respect to a distribution of tasks, albeit only in tabular settings. With rapid progress in deep RL, recent work \citep{zhang2018decoupling, franccois2019combined} has focused on building an abstract representation with a low-dimensional representation of relevant features for non-stationary settings. Learning task agnostic state abstractions has also been accomplished by identifying the causal states~\citep{zhang2019learning} in POMDPs. 

Another piece of important information that can be preserved while learning abstractions is the underlying reward and transition model, resulting in a model-irrelevance abstraction~\citep{li2006towards}. $\phi_{model}$ is a model-irrelevance abstraction if for any action $a$ and any abstract state $\hat{s}$,  $\phi_{model}(s_1)=\phi_{model}(s_2)$ implies  $R^a_{s_1} = R^a_{s_2}$ and $\sum_{s' \in \phi^{-1}_{model}(\hat{s})} P^a_{s_1, s'} = \sum_{s' \in \phi^{-1}_{model} (\hat{s})} P^a_{s_2, s'}$. In recent work, \citet{zhang2020invariant,hansen2022bisimulation,ashcraft2022structural} connect invariant causal prediction to model-irrelevance state abstractions to learn invariant representations in the Block MDP setting \citep{du2019provably}. These kinds of abstractions are even possible without explicitly modelling rewards \citep{allen2021learning}. 
Additionally, state abstractions could also be formed on the basis of value equivalence \citep{grimm2020value,sokota2021monte,cui2021control} or for improved planning \citep{curtis2022discovering}. Furthermore, in large environments with underlying Factored MDP structure, powerful abstractions have also been formed by leveraging context specific independencies that only hold for a subset of states \citep{CRADOL}.

\subsubsection{Skill Focused}
\label{sec:skillfocused}
Macro actions~\citep{hauskrecht1998hierarchical} or skills \citep{thrun1995finding} are approaches for learning that side step the requirement to make decisions at each individual time step. An MDP considers decision-making at each time-step. Humans on the other hand, are able to plan and execute tasks across multiple time scales. A semi-Markov decision process (SMDP) \citep{puterman1994markov} provides a generalized framework, in which the amount of time between two decision points is modeled as a random variable. 

Consider an agent which is in state $s$ and follows a policy $\pi_i$ in the set of available policies $\Pi$. Let's say that the transit time for the agent to enter the next state $s'$ is $\tau$ time steps; the state-transition probability from state $s$ to $s'$ could then be expressed as $p(S^{\tau} = s' | S^{0} = s, \pi_i)$. The accumulated discounted reward under the policy $\pi_i$ would then be denoted by $R_{s}^{\pi_i}$. The time for which an action persists is either a real or integer value, resulting in the SMDP model being continuous-time discrete event or discrete-time, respectively. The SMDP Bellman equations for an optimal state-value function and action-value function are then given by:
\begin{equation}
\label{eq-smdp-bellman-expectation-equation-vpi}
      v_{*}(s)  = \max_{\pi_i \in \Pi} \Bigg[ R_{s}^{\pi_i} +  \sum_{\tau = 1}^{\infty} \gamma^{\tau - 1} \sum_{s'}   p(S^{\tau} = s' | S^{0} = s, \pi_i) v_{*}(s') \Bigg]
\end{equation}
\begin{equation}
\label{eq-smdp-bellman-expectation-equation-qpi}
      q_{*}(s,\pi_i)  = R_{s}^{\pi_i} +   \sum_{\tau = 1}^{\infty} \gamma^{\tau - 1} \sum_{s'}   p(S^{\tau} = s' | S^{0} = s, \pi_i) \max_{\pi_i' \in \Pi}  q_{*}(s',\pi_i')
\end{equation}

Such an abstraction allows the agent to ignore irrelevant details and focus on learning across multiple scales of time and space. A sequence of actions forming a "macro" is one of the simplest kinds of abstraction. A macro can also be obtained as a sequence of other macros, which naturally results in a hierarchy in this architecture. The aim of hierarchical reinforcement learning is to find closed-loop policies at several levels of abstraction, also known as \textit{temporally extended actions}. Temporally extended actions are usually defined over a subset of the state space, with the primary aim being to reduce the number of steps needed for the agent to solve a task. If agents can learn abstractions, which are partial solutions to a task that could be reused for other tasks, discovering this structure in the world could potentially facilitate faster and more robust learning. 

The options framework \citep{sutton1999between} is a popular choice for temporal abstractions. An option is composed of a policy $\pi$, a termination condition $\beta: S \rightarrow [0,1]$, and an initiation set $I \subseteq S$. A \textit{Markov option} $\omega \in \Omega$ has a Markov internal policy and can be represented as a tuple $\langle I_\omega, \pi_\omega, \beta_\omega \rangle$. Options enable an MDP trajectory to be analyzed in either discrete-time transitions or SMDP-style transitions. In every state, a policy over options  $\mu : S \times \Omega \rightarrow[0,1]$ selects an option $\omega$ according to probability distribution $\mu(S_t,.)$. The option $\omega$ determines actions until $\omega$ terminates in $S_{t+k}$. Since the primitive action selection in a state $S_\tau$, between $S_t$ and $S_{t+k}$, depends not only on that time instance but also on the option $\omega$ being followed, the corresponding flat policy is considered a semi-Markov policy.

While much of the work on learning options has focused on single task learning~\citep{bacon2017option, machado2017laplacian, machado2017eigenoption, harb2018waiting, khetarpal2020options, riemer2020role,klissarov2021flexible}, work on discovering options in a multi-task context has shown some promising potential both theoretically \citep{brunskill2014pac} and empirically \citep{achiam2018variational, abstractoptions, igl2019multitask}. \citet{brunskill2014pac} derive sample complexity bounds for option discovery over a distribution of tasks. \citet{mankowitz2016adaptive} proposed a framework to learn near-optimal skills for a task, composing these skills together to enable efficient multi-task learning. 

Skills are an abstract concept that can generally be formalized as a special case of the options framework. In particular, they generally consist of partial policies, which are usually intended for reaching critical states. While options explicitly consider where to initiate a skill, which actions to subscribe to, and where to terminate the skill, work that focuses on learning skills generally consider learning a skill policy alone. Additionally, the vast majority of the work on skill learning only implicitly shows \textit{reusability} of skills in application to a low degree of non-stationarity, as opposed to explicitly concerning themselves with a full blown continual RL formulation. For instance, \citet{eysenbach2018diversity} consider a latent-conditioned policy as a skill and aim to learn a diverse set of skills in the absence of rewards to prepare for changing rewards later. With similar motivations, \citet{statecovering} propose to learn a set of skills that are state-covering rather than simply diverse. \citet{tessler2017deep} instead proposed a hierarchical, multi-skill distillation network explicitly allowing \textit{knowledge retention} and \textit{selective transfer} of skills. The Minecraft domain considered in their work enables study of non-stationarity across both rewards and transitions when agents solve multiple composite tasks.

Researchers have also considered algorithms that combine learned skills, which is one of the core objectives of continual RL. \citet{sahni2017learning, barreto2019option,lu2021reset} have focused on the natural combination of both skill and composition to allow for explicit reuse of previously acquired knowledge in the form of skills. Meanwhile, \citet{resetfree} have proposed a skill space planning framework for continual RL environments with no resets. 
Moreover, explicitly considering state and action abstraction as in \citet{abel2017toward, abel2020value} has been shown to be quite effective in theory for multi-task RL problems, albeit this theory is limited to tabular MDPs. 

\subsubsection{Goal Focused}
\label{sec:goalfocused}
Central to the mission of developing continual RL agents is the acquisition of universal knowledge, encompassing a wide variety of tasks. The meaning of a "goal" is open to interpretation and can be formalized as states the agent wants to reach, a reward the agent must achieve, or a termination point of a skill. As such, a popular strategy for decomposing complex problems in RL is to focus on goal-based reasoning. The many goal RL setting is one way to study a non-stationary setting, where changing goals are the causal source of non-stationarity in rewards. In some cases, one might even consider non-stationary settings, where generalization across goals facilitates changes in both transition dynamics and rewards over time. The earliest work on goal-based RL dates back to \citep{kaelbling1993learning}. Due to the generic nature of goals in RL, researchers have leveraged them in a variety of ways. One popular formulation that is of particular interest in the context of continual RL, is to design algorithms that generalize over goals rather than some other notion of tasks.

An ambitious approach then is to discover general purpose goals without any reward signal as in unsupervised RL \citep{jin2020reward,wang2020reward,zhang2021reward}. \citet{gregor2016variational, hausman2018learning, eysenbach2018diversity,touati2021learning} leverage principles of empowerment and discover goals based on information theoretic objectives, learning to represent the intrinsic control space of an agent in order to facilitate learning about more than a single task. \citet{achiam2018variational} relate variational discovery of options to variational auto-encoders and pose it as an optimization problem. \citet{veeriah2018many} propose a deep RL adaptation of the work of \citet{kaelbling1993learning}. Here a goal is defined as a desired raw-pixel observation, demonstrating empirical benefits in unsupervised learning without a reward signal %
for improved agent performance on downstream tasks for which goals can be treated as auxiliary tasks. \citet{andreas2017modular, oh2017zero} assume known information about high-level structural relationships among tasks and similarity between different subtasks, respectively, to enable fast learning across tasks. Meanwhile, \citet{shu2017hierarchical,geishauser2022dynamic} employ weak supervision from humans to define what skills should be learned in the form of language instructions. \citet{florensa2017automatic, chenlearning} extract %
transferable goals for a family of related tasks in multi-task RL without any assumptions about prior knowledge. 

Using a deep neural network as a value function approximator, albeit conditioned on goals, \citet{schaul2015universal} allows reasoning for a multitude of tasks. Leveraging such a universal approximator with off-policy learning from many parallel streams of experience in a continual learning fashion is very beneficial in solving tasks with deep dependencies \citep{mankowitz2018unicorn}. Moreover, \citet{zhu2019continual} leverage an unsupervised diversity exploration method to address the problem of catastrophic forgetting by learning task-specific skills similar to \citet{achiam2018variational}, alongside an adversarial self-correction mechanism to learn knowledge by exploiting past experience. Another important area of research builds off goal conditioned value functions and policies while leveraging the concept of hindsight to learn off-policy about goals that were achieved, even if they did not match the original goal of the policy \citep{HER,HPG,li2020generalized,kim2021goal,moro2022goal}. This kind of approach has been shown to greatly improve sample efficiency in sparse reward environments. Additionally, the concept of hindsight has been extended to both hierarchical \citep{levy} and modular \citep{curious} RL frameworks. 

\subsubsection{Auxiliary Tasks Focused}
\label{sec:auxiliarytaskfocused}
One of the vital requirements of a continual RL agent is to learn representations, which capture task-agnostic underlying dynamics of the world. A common approach to enable learning about more than a specific task is to augment the loss function with auxiliary losses to provide denser training signals in RL \citep{li2015recurrent, lample2017playing}. 
The vast majority of work on this topic has considered hand engineered auxiliary tasks, such as inferring the depth map from the RGB observation, detection of loop closures \citep{mirowski2016learning}, pixel control \citep{jaderberg2016reinforcement, hessel2019multi}, reward prediction \citep{jaderberg2016reinforcement}, inverse dynamics prediction \citep{shelhamer2016loss}, and latent observation prediction \citep{guo2020bootstrap}. 

A natural research direction that follows is to understand how agents can discover useful and general purpose auxiliary tasks that may ease our agent's ability to learn tasks of interest. Recent work \citep{veeriah2019discovery} addresses discovering useful auxiliary tasks in the form of the \textit{question functions} of a General Value Function (GVF) \citep{sutton2017horde}. A GVF question is a tuple $\langle \pi, c, \gamma \rangle$, conditioned on environment interaction history $h \in H$, composed of a policy $\pi: H \times A \rightarrow [0, \infty)$, cumulant $c: H \rightarrow \mathbb{R}$, and termination function $\gamma: H \rightarrow [0,1]$. The answer to a GVF question is defined as the value function, $v: H \rightarrow \mathbb{R}$, which gives the expected cumulative discounted cumulant from any history defined as:
\begin{equation}
    v(h_t) = E[c(H_{t+1}) + \gamma(H_{t+1}) v(H_{t+1}) | H_t=h_t, A_{t+1} \sim \pi(.|h_t)]
\end{equation}

Almost none of the aforementioned works explicitly consider continual RL in its entirety, and much of the advances in deep RL are yet to be fully explored in continual non-stationary environments. Moreover, there might be implications of misalignment between auxiliary tasks and the main task at hand \citep{bellemare2019geometric}. Investigating and incorporating more principled ways of discovering auxiliary learning signals is an interesting direction, particularly in the context of continual RL.

\subsection{Learning to Learn}

In this section, we will highlight applications of learning to learn or meta-learning in settings related to multi-task, lifelong, and continual RL. We will focus on three primary kinds of meta-learning. First, we will discuss approaches that learn about unknown parts of the state space. Next, we will discuss approaches that learn to improve their own ability to adapt and learn. Finally, we will highlight algorithms that learn how to explore and model curiosity driven behavior. 

\subsubsection{Context Detection}
\label{sec:contextdetection}
Very little of the non-stationarity people experience in their lives is from literal changes to the physics of the world. In fact, as mentioned in Sec. \ref{sec:taskdefinition}, the vast majority of what makes the world non-stationary is the changing behaviors of the many other agents in the world for reasons we do not understand. However, even a stationary multi-agent environment is non-stationary from the perspective of any single agent when the other agents are also constantly learning and adapting. As such, it is common to consider a problem formulation of continual RL in which an underlying MDP is assumed to be stationary but with partially observable dynamics, namely an influential \textit{task state}, which cannot be determined sufficiently from a single observation. Note that this corresponds to the partially observable view of tasks that we mention in Sec. \ref{sec:taskdefinition}. This \textit{task state} may have non-stationary characteristics, and it is necessary to infer it correctly in order to act optimally based on observations. For a real-world example, we can consider inventory management. Here, optimal behavior is dependent on the number of orders at the next step, which can be considered to be based on a not directly observed and constantly changing \textit{task state} i.e. customer demand.  

\begin{figure}[h]
  \begin{center}
  \centerline{\includegraphics[width=1.1\linewidth]{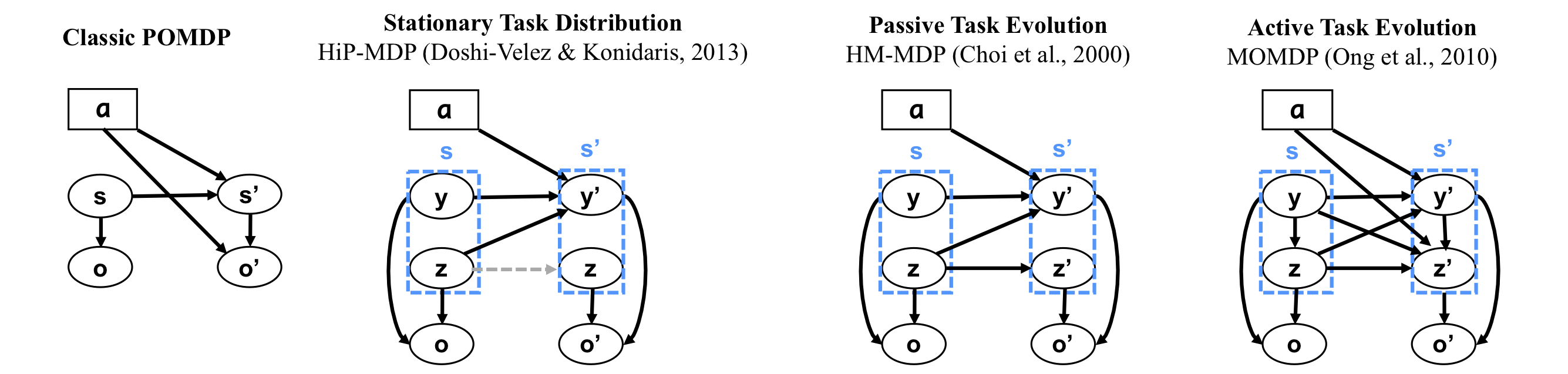}}
   \caption{\textbf{Useful Classes of Assumptions for POMDPs}: We include from left to right a classic POMDP, a HiP-MDP with stationary \textit{task state} $z$, a HM-MDP with passive Markovian \textit{task state} evolution, and a MOMDP with an active Markovian \textit{task state} evolution.}
   \label{fig:pomdpformalisms}
   \end{center}
   \vspace{-15pt}
\end{figure}

In Figure \ref{fig:pomdpformalisms} we highlight a spectrum of assumptions about \textit{task state} evolution in the literature and their connection to standard assumptions in POMDP settings. In this framework, it is generally assumed that the full state $s$ of the environment is composed of a concatenation between a within task physical state that directly produces observations $y$ and an unobserved \textit{task state} $z$. One popular formulation, called Hidden Parameter MDPs (HiP-MDPs) \citep{doshi2013hidden}, models the environment as a single fixed unknown task that may, for example, be drawn from a stationary distribution over tasks at the beginning of each episode. While it is common to explore settings with a stationary distribution of tasks, there are straightforward assumptions that can be made to help better model non-stationary dynamics over tasks. For example, Hidden Mode MDPs (HM-MDPs) \citep{hmMDP} consider this formulation for \textit{task states} that evolve independently of the agent's behavior. As seen in Figure \ref{fig:pomdpformalisms}, the task evolution of a HM-MDP is assumed to follow the behavior of a simple Markov chain $p(z'|z)$. 

A more realistic assumption in some cases may be that the tasks change at irregular and extended intervals as in the SMDP formalism \citep{hadoux2014solving}. For example, the recent work of \citep{xie2020deep} considers a formulation with Markovian transitions between tasks that change between episodes but stay constant within an episode. Alternatively, we can consider Mixed Observability MDPs (MOMDPs) \citep{MOMDP}, which are a more general formulation than passive settings, where the agent itself can potentially impact the way that the \textit{task state} evolves. In this case, the task evolution follows some stochastic function $p(z'|y,a,z)$. 

Work on context detection (i.e. discovering \textit{task states}) has often assumed access to a change point oracle. This has a very practical benefit, as it limits the length of applicable interaction history. Current work also typically assumes a stationary \textit{task state} distribution for \textit{meta-learning} \citep{doshi2013hidden,wang2016learning,duan2016rl,zintgraf2018fast,pearl,varibad,humplik2019meta,fakoor2019meta,ortega2019meta,perez2020generalized,dorfman2021offline,rigter2021risk,sodhani2021multi,fu2021towards}. However, this framework has also has been readily applicable to more challenging multi-agent learning settings \citep{da2006dealing,amato2013scalable,marinescu2017prediction,vezhnevets2019options}. These approaches have even recently been extended to learn encodings for policies themselves based on limited environment interaction \citep{harb2020policy,raileanu2020fast}. Overall, context detection is a promising approach for learning about task relatedness, which we are likely to see applied to increasingly challenging continual RL settings in the future. 

\textbf{Changepoint Detection:} As discussed previously, one core issue in addressing settings with evolving \textit{task states} is being able to detect change points or boundaries between significant switches without an oracle as in \citep{da2006dealing,rosman2012multitask,hadoux2014sequential,nonstationaryrlenvs,li2019context,kessler2022same,luo2022adapt}. However, these approaches generally tend to be reactive to a changing distribution rather than proactive about anticipated changes in the future. This kind of approach generally can converge to the optimal policy eventually for each task, but may not be able to exploit cross task dependencies effectively to improve sample efficiency.
 
\textbf{Bayesian RL:} On the other hand, a more ambitious approach would be to try to learn a belief of the unobserved \textit{task state} directly from the history of environment interactions \citep{li2009multi,hernandez2017exploration,majeed2018q}. Bayesian Adaptive MDPs \citep{martin1967bayesian,duff2002optimal} take this a step further by directly modeling uncertainty in the space of \textit{task states}. Thus in Bayesian RL, we generally seek to find a \textit{Bayesian Optimal} policy. This is a policy that acts optimally over time commensurate with its uncertainty about the current task.

\subsubsection{Learning to Adapt}
\label{sec:learningtoadapt}
A key requirement of continual RL is to acquire new capabilities in a sample efficient manner. Meta-learning is a data driven approach for improving an agent's learning efficiency. The agent attempts to learn to alter its own optimization process based on historical successes and failures of learning. To the extent that these modifications to an agent's learning algorithm generalize into the future, meta-learning should provide an inductive bias to learning that improves an agent's sample efficiency in acquiring new behaviors. Self-modifying policies \citep{SMP} provide a general framework for meta-learning in continuing RL environments. One early promising example was the Success Story Algorithm (SSA) \citep{SSA}, which leveraged backtracking for improvements that maximize a long-term average reward per step objective. SSA was also successfully applied to handle more complex multi-agent learning settings \citep{SMPMARL}. Self-modifying policies represent a very ambitious form of meta-learning, which has unfortunately not yet been scaled in application to deep RL with modern neural networks. 

\begin{figure}[h]
  \begin{center}
  \centerline{\includegraphics[width=1.0\linewidth]{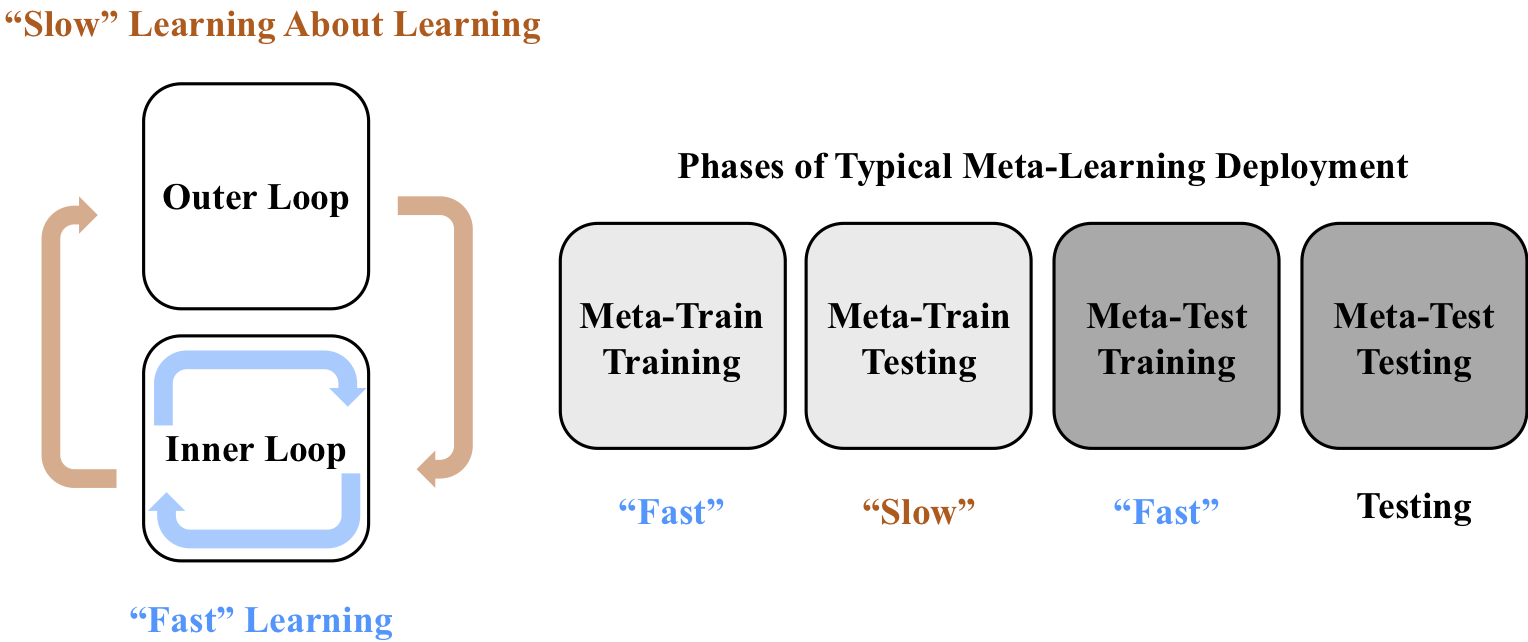}}
   \caption{\textbf{Phases of Meta-Training \& Meta-Testing}. Left: we depict the \textit{inner loop} learning process and the \textit{outer loop} learning process that aims to improve it. Right: we detail the phases of typical meta-learning, including which learning process is used in each phase. }
   \label{fig:metaphases}
   \end{center}
\end{figure}

On the other hand, there has been significant adoption of a less ambitious form of meta-optimization in the deep RL literature, which relies on a notion of what is called \textit{meta-training} and \textit{meta-testing}. We highlight some of the key details of this framework in Figure \ref{fig:metaphases}. As depicted in the left diagram, the recursive process of learning to improve an agent's own learning is decomposed into two separate learning processes called the \textit{inner loop} and \textit{outer loop} \citep{bengio1990learning,bengio1992optimization,schmidhuber1992learning}. The \textit{inner loop} is responsible for fast time-scale learning and the \textit{outer loop} is responsible for the slower process of learning about learning. For example, in the popular Model Agnostic Meta-Learning (MAML) framework \citep{MAML}, the \textit{inner loop} is standard gradient based learning and the \textit{outer loop} computes a gradient to improve the performance of the \textit{inner loop} learning process. The \textit{meta-training} process consists of \textit{meta-updates} where we learn using the \textit{inner loop} learning process on \textit{meta-train training} experiences and based on this learning, take an \textit{outer loop} learning step with the purpose of making the \textit{inner loop} more effective on \textit{meta-train testing} experiences. After we have completed \textit{meta-training}, agents are deployed in a \textit{meta-testing} setting. In this setting, we evaluate the improvements to the \textit{inner loop} learning made in \textit{meta-training} by providing new \textit{meta-test training} experiences for \textit{inner loop} learning and comparing performance on held out \textit{meta-test testing} experiences. As such, because of the disconnect between the \textit{meta-training} and \textit{meta-testing} phases of learning, these approaches can optimize for the evaluation objective despite approximating the recursive process of learning to learn by a process with two distinct steps. Indeed, this phased learning setting can be adopted for continual learning deployment at \textit{meta-test} time by training the \textit{inner loop} over a sequence of tasks \citep{Javed2019Meta,spigler2019meta,beaulieu2020learning,caccia2020online,metamapg,co2021accelerating}.

There have been a number of recent improvements on this topic of meta-optimization in multi-task RL \citep{Reptile,kim2018bayesian,rothfuss2018promp,leap,nagabandi2018deep,mendonca2019guided,finn2019online,lin2020model,berseth2021comps,co2021evolving,kirsch2022introducing,wan2022hindsight,melo2022transformers,nam2022skill} and multi-agent RL \citep{al2017continuous,lola,loladice,metamapg}. Moreover, another group of recent approaches focuses on learning a meta-critic, which explicitly guides updates to the agent's policy rather than simply guiding its actions \citep{sung2017learning,xu2019learning,harb2020policy}. While most of this work has been confined to the \textit{meta-training} and \textit{meta-testing} setting, some recent approaches have taken a first step towards applying these ideas of meta-optimization \citep{MER} and meta-critics \citep{zhou2020online,flennerhag2021bootstrapped} in a true continual RL setting closer to the spirit of self-modifying policies. For example, an approach by \citet{chandak} leverages the approximate performance of a proposed policy on the time series of past episodes to forecast its expected performance into the future and improve an agent's ability to adapt in the presence of smooth and passive non-stationarity. Additionally, a related set of approaches leverage the idea of online cross-validation \citep{sutton1992adapting}, which can be deployed in episodic environments leveraging successive episodes for \textit{meta-training} and \textit{meta-testing} \citep{MetaGrad,zahavy2020self,xu2020meta}. As these meta-learning techniques become less and less reliant on leveraging explicit training phases to segment learning, meta-learning should be able to make an impact for an even wider variety of continual RL use cases moving forward. 

\subsubsection{Learning to Explore}
\label{sec:learningtoexplore}
Another core problem of continual RL that meta-learning can potentially help with is exploration. Meta-learning has been repeatedly used in the recent literature to learn functions for intrinsic motivation and improved exploration \citep{baranes2009r,zheng2018learning,stadie2018some,xu2018learning,houthooft2018evolved,yang2019norml,zou2019reward,zheng2019can}. There are also a number of successful approaches that consider the concept of artificial curiosity. This is generally achieved by defining a heuristic based on some measurement of surprise, information gain, compression, or empowerment \citep{schmidhuber1990making,schmidhuber1991curious,klyubin2005empowerment,kaplan2006curiosity,schmidhuber2008driven,frank2014curiosity,mohamed2015variational,bellemare2016unifying,houthooft2016vime,pathak2017curiosity,karl2017unsupervised,achiam2017surprise,burda2018large,shyam2019model,liu2019competitive,ecoffet2019go,sekar2020planning,steinparz2022reactive,berseth2020smirl}.

These approaches can even be useful for learning in a single stationary environment, especially one with sparse rewards. However, there is a related curiosity problem unique to continual RL in cases where you have some level of agency over the order in which tasks are encountered. For example, the PowerPlay framework \citep{powerplay} is a methodology for deciding on the optimal task to solve next in a lifelong learning setting. There has been some limited recent work on deciding the next task to learn on in deep multi-task RL \citep{sharma2017learning}, and on learning to elicit valuable information to learn from in non-stationary multi-agent settings \citep{shu2018interactive}. However, work that considers an active and agent driven setting for exploring tasks in continual RL still remains scarce to date. This will be an interesting area to keep an eye on as it is a natural fit for RL and may be an exciting focus for future research.

\section{On Evaluation of Continual RL Agents: Benchmarks \& Metrics}
\label{sec:evaluation}
The proper procedure and protocol for evaluating continually learning agents remains an open research question. While there has been tremendous progress in the field chasing state-of-the-art results on widely acknowledged benchmarks \citep{bellemare2013arcade, brockman2016openai} and achieving super human performance \citep{silver2018general}, it is not immediately clear if the aforementioned benchmarks have the sufficient characteristics of desired environments for continual RL. \citet{schaul2018barbados} discuss a long list of open problems in continual RL. We expand on their categorization of {\em benchmarks} and {\em metrics}, presenting a discussion on the evaluation of continual RL agents. 

\subsection{Benchmarks} 
Arguably, one of the primary roadblocks in the study and rapid progress of continual RL has been the lack of well suited environments to evaluate agents in this setting. Existing and widely used benchmarks, such as classical small MDP environments (e.g. Mountain Car, Cartpole, and Taxi), OpenAI's Gym~\citep{brockman2016openai}, and the Arcade Learning Environment (ALE)~\citep{bellemare2013arcade}, have been instrumental to the progress made in RL over the years. 

Specific environments have only allowed us to study and measure agents with respect to specific dimensions. For instance, Taxi~\citep{dietterich2000hierarchical} and four rooms~\citep{sutton1999between} have exploitable structure and have been thoroughly used to study abstractions. Similarly, Mujoco and the DM-control suite were designed for continuous control, ALE for deep RL with image processing, and DeepMind Lab~\citep{beattie2016deepmind} for rich 3D navigation tasks. For more complex agents that can play long-horizon, strategy games accompanied with rich visual observations, several algorithms have also explored Minecraft~\citep{duncan2011minecraft} and VizDoom~\citep{kempka2016vizdoom}. %

Much of the work related to continual RL has hand engineered customization to the aforementioned environments to facilitate measurements for continual learning performance. In fact, often times, researchers studying continual RL end up designing tasks suitable for the specific questions they are trying to address. This might result in inherent bias in the design of experiments and tasks, which can potentially lead to unintended consequences. For instance, due to inherent determinism in some of these domains (e.g. ALE), agents have been shown to often resort to memorization of state-action sequences as opposed to achieving true generalization~\citep{machado2018revisiting}. 

Generating non-stationarity in environments in a principled way is vital for understanding the changes in learning over time. \citet{henderson2017benchmark} takes a step in this direction for multi-task RL and proposed a benchmark that parameterizes different variants of OpenAI Gym tasks, making it easier to generate novel unseen variants by modifying specific internal environment parameters in order to capture variation in transition and reward dynamics. Additionally, in recent years, a lot of progress has been made on developing benchmarks to train and test RL agents to better highlight generalization abilities. Benchmarks such as Coinrun~\citep{cobbe2019quantifying}, which is part of the larger Procgen~\citep{cobbe2019leveraging} suite of games, leverage procedural generation to create a large set of train and test environments with subtle differences. %
Additionally, \citet{osband2019behaviour} proposed bsuite, a framework with a simple set of environments aimed at specifically measuring different capabilities of an RL algorithm such as exploration, credit-assignment, and memory. Furthermore, \citet{yu2020meta} proposed Meta-world, a benchmark of 50 unique continuous control tasks for training and testing RL agents where each task can be associated with a number of environment configuration parameters.

The recent advancement of RL algorithms is in large part due to the emergence of these environments with an increased focus on testing generalization. However, most environments require a clear distinction between train-test boundaries and are dependent on a well defined notion of tasks, with limitations on how and when non-stationarity is introduced. As a result, there is still a significant need to make standard benchmarks that are tailor made for continual RL. In particular, we need benchmarks that provide rich streams of data that are configurable for varying degrees of complexity. 

A potentially promising direction would be to develop continual RL domains that allow for a range of non-stationary settings, as discussed in our taxonomy of formalisms (see Sec.~\ref{sec:taxonomyformalism}). Additionally, desirable characteristics of such benchmarks would include the ability to: 1) train in a progressive and incremental fashion, 2) facilitate discovery and composition of skills, 3) facilitate understanding of real-world dynamics such as physical rules governing the world, and 4) learn causal relationships including affordances associated with objects. \citet{CausalWorld} proposed CausalWorld, which is a promising benchmark fulfilling many of these criteria in application to robotic manipulation tasks. Furthermore, research suggests that human intelligence is grounded in learning through embodiment~\citep{kiefer2012embodiment, KielaBVC16, lake2017ingredients, bisk2020experience}. Embodied experience grounds learning in intuitive physics and causal reasoning~\citep{lake2017ingredients}. \citet{khetarpal2018environments} discusses similar recommendations for moving towards environments for lifelong agents and also highlights the benefits of embodied cognition.

Jelly Bean World~\citep{platanios2020jelly} was introduced as a test-bed for never-ending learning. In particular, Jelly Bean World supports a variety of non-stationary environment configurations including multi-task, multi-agent, multi-modal, and curriculum learning settings. Although designed for never-ending learning, the framework has considerable overlap with lifelong/continual RL, making it suitable for evaluating continual learning in its purest form.\footnote{Never-ending learning, as posed in the Jelly Bean World environment, suggests that the notion of a task or subtask naturally emerges and the distinction between tasks is not always as sharp as in literature on lifelong learning or continual learning. Never-ending learning can also be formulated within our taxonomy of formalisms, where the nature of the non-stationarity allows for these lines to be blurry.}

More recent advances targeted at evaluating continual learning capabilities have addressed some of the aforementioned desiderata of CRL benchmarks. \citet{nekoei2021continuous} introduced a Hanabi based multi-agent lifelong learning testbed and evaluated many multi-agent RL approaches. \citet{powers2021cora} introduced CORA, a unification of benchmarks, metrics, and baselines incorporating ALE, NetHack, Procgen and CHORES. \citet{johnson2022l2explorer} presented a highly configurable, Unity-based environment for testing continual and lifelong learning systems. \cite{goel2021novelgridworlds} on the other hand, is a benchmark focused on exploring continual sources of novelty.

In the context of non-stationarity in real-world applications, researchers have also explored large scale recommender systems~\citep{chandak2019learning} and diabetes treatment~\citep{chandak2020safe} as case studies for when the nature of RL problems have inbuilt non-stationarity to account for.

\subsection{Metrics} 
The longstanding tradition in reinforcement learning has been to measure an agent's performance by recording its average expected accumulated rewards over time \citep{Sutton98}. While reward is a good quantitative measure of an agent's performance, measuring expected accumulated returns by the agent may not suffice for fully understanding the abilities of a continually learning agent. The choice of metrics in RL is often tightly coupled with the choice of the problem formulation. Consider an RL algorithm designed to solve many sequential tasks that are related to each other through proper skill discovery (see Sec.~\ref{sec:skillfocused}). While accumulated returns is a strong primary indicator of learning performance, it does not give us any insight into the robustness of skills learned or if the skills learned in one task are leveraged across the lifetime of an agent. Furthermore, for a continual learner, it is vital to measure core metrics, for example, forward and backward transfer (or interference), skill reusability, and skill composition.

A promising approach is to incorporate the idea of \textit{auxiliary metrics} \citep{schaul2018barbados} to measure an agent's intelligence with {\em probe questions} that function in a similar way to auxiliary losses. Incorporating auxiliary tasks (see Sec.~\ref{sec:auxiliarytaskfocused}) has been shown to improve the representations learned by an agent. Moreover, auxiliary evaluation metrics can further our understanding of an agent's abilities. %
Specifically, the core desired capabilities of a continual learner can be tested with probe questions.

\subsection{Towards Broader Evaluation Criteria for Continual RL}
\begin{figure}[h]
\vspace{-2pt}
  \begin{center}
  \centerline{\includegraphics[width=1.05\linewidth]{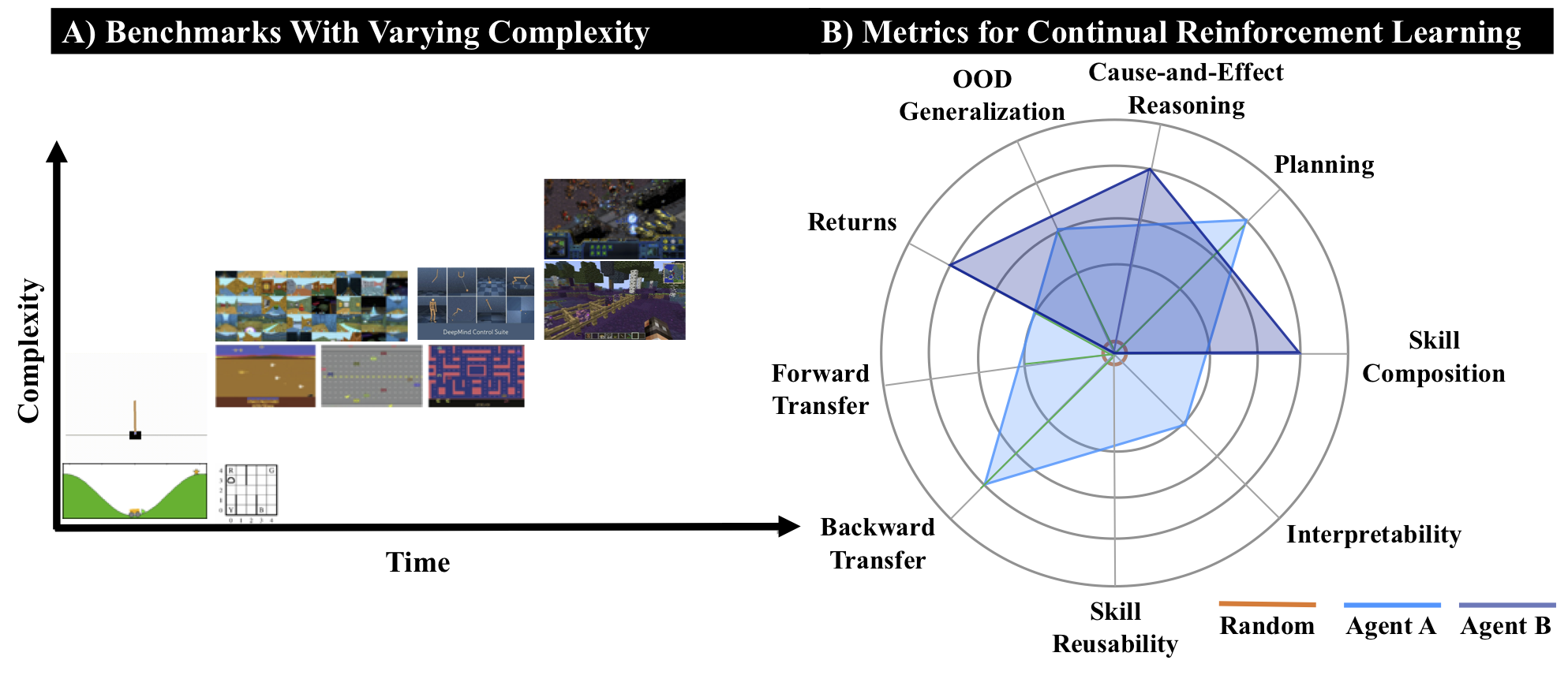}}
  \vspace{-4mm}
   \caption{\textbf{Evaluating Continual Reinforcement Learning Agents}. A) Depicts the evolution of domains and benchmarks over time commonly used in RL. B) Depicts key metrics for evaluating continual RL agents in the style of bsuite. Such a framework should also offer a knob controlling the degree and nature of non-stationarity that agents experience (see Figure \ref{fig:taxonomy_formalisms}). For a given degree of non-stationarity, a set of carefully designed experiments to test different \textit{probe questions} would help foster more rapid progress in the field.}
   \label{fig:unifiedcrlframework}
   \end{center}
\end{figure}

At the intersection of metrics (what to measure) and domains (how to measure)\footnote{We acknowledge that evaluation of deep RL agents faces several challenges pertaining to reproducibility (see ~\citep{henderson2017deep, khetarpal2018re}). This is even more reason for the community to move towards standardized evaluation benchmarks for continual RL.} we recommend that bsuite \citep{osband2019behaviour} is a promising example of the type of framework needed for training and evaluating agents in a continual fashion. 
In Figure~\ref{fig:unifiedcrlframework} we highlight some important evaluation criteria to consider to better understand the performance of continual RL agents.
For a given degree and nature of non-stationarity (see Sec.~\ref{fig:taxonomy_formalisms}), researchers should generate a set of carefully designed experiments with a carefully chosen complexity to train and evaluate continual RL agents. Ideally, proper empirical analysis would result in a measure of the behavior along different dimensions of probe-metrics as shown in the Fig.~\ref{fig:unifiedcrlframework}.

To this end, it is important to consider the following capabilities as probe questions (i.e. auxiliary metrics) in addition to measuring the accumulated returns over time.
\begin{enumerate}
    \item \textit{Catastrophic Forgetting (Forward and Backward Transfer)}: It is desired for our agents to be able to effectively use previously acquired knowledge in new related situations that they might encounter (i.e. forward transfer). Moreover, if an agent has seen a situation before and encounters a similar experience, it should be able to perform backward transfer to improve its previously learned capabilities. When an agent's current learning greatly interferes with its ability on previous experiences, the agent is experiencing catastrophic forgetting. %
    \item \textit{Skill Reusability}:  Just as humans acquire skills and build on them to solve increasingly complex tasks, a continual learning agent must be able to reuse previously learned skills in new unseen situations. This is an important ability, especially when new skills can be created on-the-fly in new situations. Measuring skill reusability is challenging because this might include adjustments and tweaks to the previously learned representations or explicit skills. Much of the existing work uses qualitative analysis to measure reusability or adaptability. See Sec.~\ref{sec:skillfocused} for a detailed discussion on skill focused approaches.
    \item \textit{Qualitative Analysis (Interpretability)}: Understanding and interpreting different aspects of behavior through qualitative analysis is undervalued in the field, while the heavy emphasis is on improving scores and learning curves. We posit that qualitative analysis that provides clarity about the type of representations learned, the kind of behaviors acquired, the landscape of the value functions and policy, the changes made to previously learned knowledge, and model predictions will all be key to furthering our understanding of continual RL methods. %
    \item \textit{Skill Composition}: A common desiderata for continual RL agents is to effectively leverage shared structure over the data that an agent sees. See Sec.~\ref{sec:sharedstructure} for a detailed discussion. Composing previously learned behaviors to perform new ones is an important capability to consider, as this enables agents to exploit what was learned before with greater efficacy. %
    \item \textit{Planning}: A lifelong learning agent must be able to effectively plan for the future by leveraging its acquired knowledge. Planning can be explicitly measured by evaluating an agent's explicit or implicit plans over time on different tasks. For instance, if an agent is building models of the world, measuring the value function through approximate dynamic programming can give an estimate of how well the agent can plan. Such a procedure should be adaptable for planning in both observation spaces and latent representation spaces, including single time-step and multi-time-steps extrapolation.
    \item \textit{Cause and Effect Reasoning}: Designing probes for cause-and-effect analysis will allow a quantitative measure of how well continual RL agents are learning the underlying rules and objects of an environment. One concrete way to measure this is to design object-oriented probes associated with well grounded perturbations on objects to model \textit{interventions} and test an agent's causal understanding. For example, this kind of evaluation is possible leveraging CausalWorld \citep{CausalWorld}. 
    \item \textit{Out of Distribution (OOD) Generalization}: Another probe-metric to test agents on over the course of their lifetime would be to measure an agent's performance when we situate it in environments that lie outside of its prior training distribution. In these cases, it would be interesting to evaluate an agent's generalization performance by measuring both zero-shot expected returns and its sample complexity when learning new capabilities.
\end{enumerate}

\section{Looking Forward}
\label{sec:lookingforward}

In this section we conclude our survey by looking forward at the frontiers of continual RL research. Specifically, we first discuss potential connections between continual RL research and findings in the neuroscience community. We then discuss challenges and open problems that continual RL research will have to address in order to make progress. 

\subsection{Bridging the Gap Between Continual RL and Neuroscience}
\label{sec:neuro}

Findings in the neuroscience literature have often served as a guiding light towards developing human-like continual learners. How humans have the innate ability to perform long-term decision making without a task reward, or rather even delayed sparse rewards, remains largely a mystery. In fact, it is unclear even what utility~\citep{samuelson1937note} humans optimize for. Nonetheless, there has still been interesting work at the intersection of neuroscience and RL. For example, \citet{HASSABIS2017245} provides a survey on neuroscience inspired AI, and \citet{niv2009reinforcement} details evidence that RL happens in the human brain. We now briefly discuss some areas of interest to continual RL where it may be potentially useful to draw insights from research on the human mind. 

\paragraph{Balancing Stability and Plasticity.} The tension between prioritizing recent experiences and past experiences, also known as the \textit{stability-plasticity dilemma}, is often encountered when training neural networks and has also been demonstrated in the human brain~\citep{StabilityPlasticity}. This additionally closely ties in with the inherent problem of credit assignment in RL.  Humans have somewhat seamless mechanisms in place to assign credit, even for events which happen far after the relevant actions that caused them. Inspired by these findings in neuroscience, the recent work of~\citet{hung2019optimizing} introduces a temporal value transport algorithm, where the agent uses specific memories to credit past actions. 

The neuroscience literature can give us more insight into the nature of task interference that humans experience. For example, \citet{detre2013moderate} demonstrate that human memories that are highly activated during continual learning are less likely to be forgotten later. Additionally, \citet{sagiv2020efficiency} theorize that the inability of humans to effectively perform multiple tasks at the same time results from a trade-off related to the human tendency to share network architectures between tasks in order to enable quicker learning. While AI agents suffer a great deal from  {\em catastrophic forgetting}, evidence in the literature~\citep{mcclelland1995} suggests that the mammalian brain may avoid catastrophic forgetting by protecting previously acquired knowledge in neocortical circuits through interaction with the hippocampus. Leveraging these findings~\citet{kirkpatrick2017overcoming} discuss how continual learning in the neocortex relies on task-specific synaptic consolidation, whereby knowledge is durably encoded by rendering a proportion of synapses less plastic and therefore stable over long timescales. Finally, \citet{ajemian} demonstrate the role that noisy computations in the human brain may have in avoiding catastrophic forgetting, and \citet{dohare2021continual} also demonstrate that noisy computations can improve plasticity. 

\paragraph{Nature of Human Rewards.} One place where inspiration from neuroscience may be particularly helpful for designing AI systems is in understanding more about the origins of reward. It has been suggested that an average reward per step objective is more consistent with human studies than the discounted cumulative reward objective that has become more popular for RL research \citep{daw2000behavioral}. 
There is also evidence that humans prefer to acquire more knowledge even when it lacks predictive value \citep{niv2011value}. Indeed, there has been a significant body of research trying to understand the role of the dopamine system as a reward signal for humans \citep{samson2010computational,starkweather2017dopamine}. It has also been theorized that the dopamine system is directly responsible for a kind of meta-learning by training the prefrontal cortex, which then can function as its own standalone learning system \citep{wang2018prefrontal}. Moreover, it has been demonstrated that stress can play a big factor in selectively modulating the reward signal for humans \citep{berghorst2013acute}. Finally, it has been recently suggested that many sophisticated aspects of human learning cannot be explained by RL, and that these aspects of behavior may be supported by the brain’s executive functions \citep{rmus2020role}.

\paragraph{Leveraging Memory.} The neuroscience literature related to the interplay between learning and memory formation is another potential source of insight for building better continual RL agents. For example, \citet{collins2012much} study how RL and working memory complement each other in the human brain. Additionally, there is strong evidence in the literature of hippocampal replay facilitating the model-based planning process for human RL \citep{mattar2018prioritized,momennejad2018offline,vikbladh2019hippocampal,momennejad2020learning}. Interestingly, the close connection between experience replay and planning was also recently highlighted in the RL literature \citep{pan2018organizing,eysenbach2019search}. Moreover, it has been suggested that humans perform a form of pseudo-replay or pseudo-rehearsals that is likely particularly active and useful for consolidation of knowledge during human sleep \citep{Robins95,Robins96,Frean1999}. \citet{schuck2019sequential} also suggest that hippocampal replay may be important for building representations of complex, abstract tasks elsewhere in the brain. Likewise, it has been suggested that the hippocampus may be central to the superior value generalization capabilities that humans demonstrate \citep{wimmer2012generalization}. In fact, there is evidence that it also may be critical for computing value over complex state spaces, learning with little data, and performing long-term credit assignment \citep{gershman2017reinforcement}. Finally, it has been demonstrated that reward prediction errors also play a key role in the memory forming process \citep{jang2019positive}, which also mirrors findings in deep RL \citep{schaul2015prioritized}.  

\paragraph{Balancing Model-Based and Model-Free Learning:} There has also been significant research on the interplay between model-free and model-based learning in the human brain that could provide guidance in designing sample efficient continual RL agents. Research suggests that a combination of model-free and model-based computations are performed in the human brain \citep{glascher2010states,daw2011model,doll2012ubiquity}. In fact, \citet{langdon2018model} even highlight the influence of model-based computations rather than only model-free ones in dopamine responses. It is believed that humans primarily rely on model-based learning for dealing with tasks that have both high volatility and low noise \citep{simon2011environmental}. Additionally, model-based reasoning has been shown to prevent humans from forming habits \citep{gillan2015model}. Moreover, recent work has demonstrated that the successor representation model from the RL literature \citep{dayan1993improving} may provide a more accurate model to reflect the way humans balance their use of both model-based and model-free learning \citep{momennejad2017successor}. 

\paragraph{Exploiting Modular Structure.} \citet{doya2002multiple} proposed a modular control architecture using multiple prediction models based on the computational model of the cerebellum proposed by \citet{wolpert1998internal}. Their simulation results corroborate findings associated with fMRI data \citep{imamizu1997separated, imamizu2000human} suggesting that when a new task is introduced, many modules initially compete to learn it. However, after repetitive learning, only a subset of modules are specialized and recruited for the new task. Related work by \citet{daw2005uncertainty} proposes a Bayesian uncertainty based model for arbitrating competition among important subsystems of the human brain. Indeed, studies of the human brain are a fruitful reference point for understanding what may be good paradigms for exploiting a modular architecture in the context of continual RL. 

\paragraph{Results Corroborating Current Trends.} Finally, there is a growing body of evidence that justifies many currently popular trends in the continual RL literature based on human comparisons. 
For example, there is significant evidence in the literature that human learning is either uncertainty-aware or Bayesian in nature \citep{niv2005dopamine,fleming2017self,babayan2018belief,lowet2020distributional}. There is also significant evidence that substantial hierarchical structure is present in the human learning process \citep{botvinick2009hierarchically,gershman2009human,frank2012mechanisms,badre2012mechanisms,diuk2013divide,botvinick2014model,solway2014optimal,ribas2019subgoal}. Moreover, \citet{niv2015reinforcement} highlights the role of attention in the human brain in supporting the reinforcement learning process to address the curse of dimensionality. Interesting work by \citet{niv2019learning} also suggests that the orbitofrontal cortex may be used to represent \textit{task-states} that are deployed for decision making and learning elsewhere in the brain. It has been demonstrated that human RL attempts to identify the causal structure in the task at hand \citep{gershman2015discovering,gershman2017reinforcement}.

\subsection{Challenges and Open Problems}
Due to the lack of a concrete definition and the extremely general nature of CRL, there is still significant unexplored potential to be addressed by future research. Moreover, there remain a number of open problems which come with their own challenges. In this section, we will discuss some of these fundamental challenges in more detail.\footnote{We refer the reader to \citep{schaul2018barbados} for more discussion on open problems in continual RL.}

\paragraph{Inductive Biases.}
There are, without a doubt, many different perspectives within the AI community on how much and to what degree we should embed inductive biases in our agents. However, leveraging priors from the human learning process has enabled the field to make significant progress over time. Many real-world applications such as robot navigation and autonomous driving are tangible now only as a result of making such assumptions 

In the context of a continual RL agent, this remains an open question and of greater significance. What utility an agent optimizes for over its lifetime is not immediately clear even for human learning. \citet{sutton2019bitter} argues that the most promise lies in leveraging computation, as opposed to leveraging human knowledge and inductive biases coming from this human knowledge. ``\textit{We have to learn the bitter lesson that building in how we think we think does not work in the long run.}"~\citep{sutton2019bitter}. To this end, potential directions aligned with this process of discovery through computation include: self-tuning approaches~\citep{MetaGrad,zahavy2020self}, end-to-end skill discovery~\citep{bacon2017option}, discovering RL algorithms~\citep{kirsch2019improving,oh2020discovering}, learning what objective to learn~\citep{xu2020meta}, and other approaches moving more and more towards open-ended learning~\citep{wang2020enhanced}.

\paragraph{Task Specification.} Expressing \textit{task specification}, which often carries subtle assumptions as to how tasks are related, is often left to the interpretation of the designer. Although a task can be specified as its own MDP, this definition is very broad and thus limited in that sense. To overcome the extremely general nature of this definition, researchers have proposed several MDP variations of relevance to continual RL such as HM-MDPs, MOMDPs, Block MDPs, HiP-MDPs, and Factored MDPs. 
Moreover, where tasks and rewards actually come from is another open and somewhat philosophical problem that is a long-standing question for researchers. Indeed, hand engineering of rewards (and therefore tasks) has been under scrutiny for decades without much progress in making agents less dependent on this level of human intervention. %

\paragraph{The Agent-Environment Boundary.}
Traditionally MDPs serve as a framework for formalizing agent-environment interaction and the corresponding boundary between the two. Considering the role of other agents in the learning of real-world decision makers, the correct way to view the agent-environment boundary remains an open question of high significance. \citet{jiang2019value} and \citet{harutyunyan2020} offer fresh perspectives on this discussion. The taxonomy presented in this work mostly included literature that is concerned with the traditional view of a single decision making agent with everything else delineated as the environment. On the other hand, multi-agent formulations explicitly consider the presence of other agents and model learning with this alternative view. %
In the context of non-stationarity, the agent-environment boundary is an open concept because the space of affordances~\citep{khetarpal2020can} might evolve over time~\citep{pezzulo2016navigating} and may only emerge as a result of agent-environment interaction~\citep{gibson1977theory, heft1989affordances,chemero2003outline}.

\paragraph{Experiment Design and Evaluation.}
As we discussed earlier, studying the full non-stationary setting in its entirety is a significant challenge for continually learning agents. Designing and engineering such experiments is not only laborious, but also subject to the scale at which one can study such a setting. We have discussed possible benchmarks with a unification of domains and metrics (see Sec.~\ref{sec:evaluation}) to generate systematic experiments for training continual RL agents. However, another related open question in continual RL has to do with the nature of evaluation itself. Indeed, it seems somewhat unnatural to have separate validation or testing phases in a true continual RL setting. In some sense, the notion of separate phases for evaluation that prohibit learning can be seen as idealistic, as it is not generally possible to really test humans in such a way. 

Moreover, disconnects between training and testing settings can be problematic for optimization in the context of continual RL. For example, it is popular to test continual learning agents in the so called \textit{one pass} setting (where an agent is trained on each task in succession while never revisiting old tasks and then tested on its retention over the distribution of old tasks). This setting would be problematic for any continual RL agent without a pre-specified prior towards remembering old tasks. This is because a purely data driven agent has seen no evidence that past tasks will reoccur, and thus has no reason to maintain performance on these tasks that is conveyed to it through its rewards during training. As a result, even basic questions about how to evaluate agents can be quite tricky for continual RL and can be considered a challenge that the community still has to understand more deeply moving forward.

\paragraph{Interpreting Discovery.}
The advances in image classification tasks with the advent of large scale labeled datasets such as ImageNet~\citep{krizhevsky2012imagenet, russakovsky2015imagenet} was followed by a plethora of research on visualizing \citep{zeiler2014visualizing} and understanding~\citep{szegedy2013intriguing} the nature of deep convolutional neural networks. A large body of research in continual RL, which address {\em the discovery problem}, makes an overarching promise for a certain kind of solution (e.g. those that are general and adaptive), but most of these claims are really hard to verify. A natural technique researchers adopt is qualitative analysis, which is often subject to interpretation. To advance research towards the overarching goal of artificial general intelligence, we need to develop tools to understand and introspect our agents based on more than just rewards.

\paragraph{Learning at Scale.}
Much of the recent revolution within machine learning, and in deep learning in particular, is in large part due to advancements in hardware~\citep{hooker2020hardware}. It is arguable that continual RL agents might overcome potential issues such as forgetting if learning is performed at scale. Recently, we have seen similar findings in the natural language processing community, where language models such as GPT-3 ~\citep{brown2020language} are able to perform much better due to learning at scale. This is not to say that throwing a lot of data and compute at the problem will solve everything, nor is it the one solution we need. However, when connections to evolutionary learning in the human brain are made, we do need to really appreciate the scale at which such learning takes place. Recent work has looked at the topic of \textit{scaling laws} in deep supervised learning \citep{kaplan2020scaling,sharma2020neural,henighan2020scaling}. Exploring the relevant \textit{scaling laws} for continual RL will be a potentially fruitful topic to understand more clearly moving forward. 

\paragraph{Learning in the Presence of Other Learning Agents.}
As highlighted in Proposition \ref{prop:activemarkovgame}, a key continual RL scenario is when an agent must learn in a stationary environment that contains other agents that are also learning. The non-stationarity experienced in these domains is a key consideration of multi-agent RL (MARL). See \citet{hernandez2019survey} for a comprehensive survey on MARL. In these settings, even the correct solution concept can be difficult to define \citep{littman01friendfoe,wang02nash,bowling2004convergence,greenwald03correlated,zinkevich06cyclic,letcher2018stable,further}. That said, the existence of other agents in the environment could be a blessing if treated appropriately. For example, agents can teach other agents as they learn so that each learns to better perform tasks that are required of them \citep{torrey2013teaching,omidshafiei2019learning,kim2020learning}. Moreover, agents can learn to shape the learning of other agents and thus the nature of the non-stationarity in their environment \citep{zhang10lookahead,lola,letcher2018stable,loladice,metamapg,further}. MARL is quite far from use in industrial applications and is still largely in its infancy as a research direction. However, it is clear that developing agents capable of acting effectively in large systems with other interacting agents will be an important step towards enabling continual RL agents to integrate with our society and navigate the real world outside of simulation.

\section{Conclusion} 
In summary, we would like to highlight final considerations. While finding the right set of assumptions is an important meta challenge of continual RL, we would like to push the continual RL research community towards increasingly realistic settings. Findings about the human brain, psychology, and cognitive behaviors, including animal learning, have laid strong foundations for the field of reinforcement learning. Sec.~\ref{sec:neuro} further highlights that bridging the gap between AI and computational neuroscience has promising potential to help make rapid advancements in the field of continual RL. We should aim to address these open problems of continual RL with the aspiration to deploy agents in challenging real-world applications, where continual RL has potentially promising use cases. Supervised continual learning has seen some success in the same spirit; \citet{carlson2010toward} is one example deployed in a true never-ending continual learning fashion, albeit in a very controlled setting. More recently, Blenderbot~\citep{shuster2022blenderbot} is also a step in this direction in the context of conversational agents. While largely a theoretical field to date, continual RL marks a shift towards a more robust style of learning that, when successful, will greatly broaden the applicability of RL for real-world use cases. We hope that this survey serves as a useful resource for the continual RL research community that will help us collectively understand and eventually achieve this lofty goal.

\section*{Acknowledgements}
We would like to thank Takuya Ito and Martin Klissarov for providing valuable feedback. We would also like to thank Joelle Pineau for insightful comments on an earlier draft of the paper. We really appreciate the feedback we received from the JAIR reviewers, which helped improve the contributions of this work. Finally, we would like to give a special thank you to Anna Riemer for editing our final manuscript. We acknowledge that this work originated as a class project undertaken in the graduate-level course on \href{https://sites.google.com/site/irinarish/continuallearning}{Continual Learning: Towards "Broad" AI (IFT-6760B)} at Mila, Montreal. IR also acknowledges support from the Canada CIFAR AI Chair Program and the Canada Excellence Research Chairs Program.

\appendix 

\newpage

\section{The Relationship Between Continual Reinforcement Learning and Continual Supervised Learning} \label{app:supervisedlearning}

As noted in \cite{barto2004reinforcement}, supervised learning can be cast as a special case of RL in continuing environments. In this section, we consider a particular set of assumptions that simplify the RL framework to better match the typical supervised learning setting:

\begin{enumerate}
    \item \textbf{Deterministic Policies:} Instead of a potentially stochastic policy $a \sim \pi(s;\theta)$ with parameters $\theta$, we consider a deterministic function $\hat{y}=f(x;\theta)$ of the input $x$ where the predicted output can be interpreted as a sampled action $A_t=\hat{Y}_t$ associated with sampled input $X_t$.
    \item \textbf{Decomposed State Space:} The state space must provide enough information to compute the reward of a given action, so we consider the full state space of supervised learning to include both the input $x$ and optimal output $y^*$ i.e. $S_t=(X_t,Y_t^*)$. In this formulation, the supervised learning problem is partially observable (if $Y^*_t$ is provided the problem becomes trivial). 
    \item \textbf{Differentiable Reward Function:} In supervised learning we generally consider a differentiable loss function in lieu of a reward function i.e. $r(s,a)=-\ell(y^*,\hat{y})$.
    \item \textbf{Action Invariant Transitions:} We also assume that the transition dynamics of the environment $p(s'|s,a)$ do not depend on the agent's behavior. The data distribution is is rather drawn from the joint distribution of inputs and optimal outputs $p(x,y^*)$. 
\end{enumerate}

Following from these assumptions and considering the limit of the undiscounted problem i.e. $\gamma \rightarrow 1$ during a lifetime $T$, the continuing environment objective function from equation \ref{infinitehorizonobjective} can be extended to the supervised learning setting as: 

\begin{equation} 
\begin{split}
J_{\text{continuing}}(\theta) &= -\mathbb{E}_{p(x,y^*)} \bigg[ \sum_{k=0}^T \ell(Y^*_{t+k},\hat{Y}_{t+k}=f(X_{t+k};\theta)) \bigg] 
\end{split}
\end{equation}  

Notice that if we assume that the incoming data distribution is stationary, we can consider the linearity of expectations over random variables and bring the sum outside of the expectation:

\begin{equation} 
\begin{split}
J_{\text{stationary}}(\theta) &=  -\sum_{k=0}^T \mathbb{E}_{p(x,y^*)} \bigg[ \ell(Y^*_{t+k},\hat{Y}_{t+k}=f(X_{t+k};\theta)) \bigg] 
\end{split}
\end{equation}  

If $p(x,y^*)$ is assumed to be i.i.d., samples from this sum are unbiased, drawing an equivalence with the standard stochastic gradient descent (SGD) supervised learning objective:
\begin{equation} 
\begin{split}
J_{\text{SGD}}(\theta) &= -\mathbb{E}_{p(x,y^*)} \bigg[ \ell(Y^*_{t},\hat{Y}_{t}=f(X_{t};\theta)) \bigg] 
\end{split}
\end{equation} 

However, in continual supervised learning, we generally consider the data to be following some non-stationary pattern $p(x,y^*,t)$. Unfortunately, the linearity of expectations only applies over random variables, so it becomes clear that the SGD objective is biased towards the current experience without considering the long-term effects of parameter changes:

\begin{equation} 
\begin{split}
J_{\text{non-stationary}}(\theta) &= -\mathbb{E}_{p(x,y^*,t)} \bigg[ \sum_{k=0}^T \ell(Y^*_{t+k},\hat{Y}_{t+k}=f(X_{t+k};\theta)) \bigg] \\
&= - \mathbb{E}_{p(x,y^*,t)} \bigg[  \ell(Y^*_{t},\hat{Y}_{t}=f(X_{t};\theta)) + \sum_{k=1}^T \ell(Y^*_{t+k},\hat{Y}_{t+k}=f(X_{t+k};\theta)) \bigg] 
\end{split}
\end{equation} 

When experiences from the past data distribution are expected to re-occur in the future distribution but not the current distribution, the biased objective of SGD may thus naturally lead to catastrophic forgetting of this past knowledge. Notice that the SGD objective has originally been shown to converge in the stationary and i.i.d. setting, so outside of that setting we must use another valid framework to model learning. From this analysis, it is clear that the RL framework is general enough to capture many of the important features of continual supervised learning in non-stationary settings.

\bibliography{sec_bib}
\bibliographystyle{unsrtnat}

\end{document}